\documentclass[11pt,letterpaper]{article}
\usepackage[margin=22mm]{geometry}
\usepackage{fontspec}
\usepackage{amsmath,amssymb,graphicx,array,longtable,xurl,hyperref,placeins,fancyvrb,newunicodechar}
\newunicodechar{≈}{\ensuremath{\approx}}
\newunicodechar{≥}{\ensuremath{\geq}}
\hypersetup{hidelinks,pdfauthor={Aleksandr M. Shtraukh-Adamsky; Vladimir M. Adamsky; Irina G. Voloshina; Artemiy D. Spitsyn},pdftitle={When Does Less Information Pay?
An Exact Finite Atlas of Representation, Payoff, and Cost}}
\begin{document}

\begin{center}
{\Large\bfseries When Does Less Information Pay? An Exact Finite Atlas of Representation, Payoff, and Cost\par}
\end{center}

\textbf{Manuscript status:} Preprint, version 1.0 (not peer reviewed).

\textbf{Article license:} CC BY 4.0.

\textbf{First-author ORCID:} https://orcid.org/0009-0006-8458-8033

\textbf{Authors:} Aleksandr M. Shtraukh-Adamsky, Vladimir M. Adamsky,
Irina G. Voloshina, Artemiy D. Spitsyn

\textbf{Affiliation for all authors:} Independent researchers, Moscow, Russia

\textbf{Corresponding author:} Aleksandr M. Shtraukh-Adamsky,
\href{mailto:alogin7@yandex.ru}{alogin7@yandex.ru}

\noindent{\small Code, data, editable sources, and Russian translation: \url{https://doi.org/10.5281/zenodo.23868864}.}\par
\FloatBarrier
\section*{Abstract}

A representation that retains one bit about a four-state world can solve a task
perfectly, while another that retains 1.5 bits on average reaches only 75\% accuracy.
The difference lies in which distinctions each representation preserves. We examine
this relationship in a uniform four-state world by enumerating all admissible
representations and binary action rules. Constructions are compared both by replacing
one component while holding the others fixed and by choosing the best action rule
separately for each module. We consider three tasks, their mixtures, and four
separate hypothetical cost tariffs.

In some comparisons, the representation retaining less world information has the
higher payoff; in others, the more informative representation does. There are also
pairs with different information amounts but equal payoff—that is, equal task
accuracy. If a representation preserves all existing
distinctions and adds new ones, its best zero-cost payoff cannot decrease.
Lower-information advantages therefore depend on \emph{which} distinctions are retained.
Equal counts of opposing advantages under representation replacement follow from
coding symmetry. In broader comparisons, the shares depend on how repeated
realizations are counted. A different task set or resource price can change which
member of a fixed pair performs better.

These results are exact, complete censuses of the specified finite sets of
constructions and comparisons. The atlas provides a reproducible worked model
for examining how action constraints, counting rules, and cost assumptions
affect comparisons. It illustrates established distinctions between information
quantity and decision value rather than proposing a new general law of their relation.
\(I(W;R)\) measures retained world information, and payoff measures task accuracy.

\textbf{Keywords:} representation; mutual information; task payoff; exact enumeration;
fitness–truth debate; finite comparison; rate–distortion; computational reproducibility

\FloatBarrier
\section*{1. Introduction}

If an agent cannot retain all the available information about its environment,
which distinctions between states should it preserve? The answer depends on the
task: a representation may contain more information about the world overall,
yet fail to preserve the particular distinctions needed for the right action.

Consider four equiprobable world states, labelled by two-digit binary codes:
\(00,01,10,11\). The agent's task is to identify the first, left-hand bit.
A representation that distinguishes \(00,01\) from \(10,11\) allows error-free
answers and retains one bit of world information. Another representation merges
\(00\) with \(10\) but distinguishes the other two states. It retains 1.5 bits
on average, yet cannot always answer correctly.

In this example, one bit of information means that two of the four equiprobable
possibilities remain. This happens every time with the first representation.
With the second, half the time—in states \(01\) or \(11\)—the world state
is known exactly, providing two bits of information. The other half of the time,
the possibilities are \(00\) and \(10\), providing one bit.
The average is \((1+2)/2=1.5\) bits. The fractional value describes the
representation's average informativeness.

The first representation's advantage comes
from \textbf{which distinctions it preserves}, not from the reduction in information
itself. As shown in Section 3.2, simply removing distinctions from a given
representation cannot improve its best attainable accuracy.

The distinction between information quantity and task relevance is already central
to information-theoretic compression. The deterministic information bottleneck
balances representation entropy against information retained about a target
(Strouse \& Schwab, 2017). Tegmark and Wu (2020) study such trade-offs for binary
classification. These criteria concern information about a target, whereas our
payoff measures the accuracy of a chosen action rule; the two are not interchangeable.

Tan et al. (2022) search the space of deterministic groupings for a Pareto frontier:
representations for which neither compression nor target information can be improved
without sacrificing the other. Their method also applies to other two-objective
clustering problems, and they use exhaustive enumeration as a small-case reference.
Our purpose is different: to retain all specified pairwise comparisons, including
nonoptimal constructions, and examine the consequences of fixed versus separately
optimized action rules, repeated realizations, task mixtures, and distinct costs.
The atlas is not a new algorithm for finding optimal representations.

The motivation for this study came from Donald Hoffman's \emph{The Case Against Reality}
and the evolutionary modelling literature it discusses (Hoffman, 2019; Mark et al.,
2010; Prakash et al., 2021). A central question in this debate is whether
fitness-oriented strategies can outperform strategies described as veridical.
Addressing it requires distinguishing representation, information, payoff, and
biological fitness. A representation specifies which world states an agent
distinguishes. Information quantifies statistical dependence between the
representation and the world. Payoff scores the success of selected actions in
a given task. Relating this payoff to reproductive success requires a separate
model.

Previous studies show why the answer cannot be separated from the comparison
conditions. Advantages for simpler or interface-like strategies have been found
under particular payoff and cost assumptions (Mark et al., 2010); a general
mathematical result has also been obtained for specified strategy classes
(Prakash et al., 2021). Other factors may matter, however: information that is
useful only when several input signals are combined (Martínez, 2019),
environmental change (Angelucci et al., 2021; Mattes, 2026), and the need to solve
multiple independent tasks using one representation that does not adjust to
the current goal (Berke et al., 2022). Preserving relations between world states
is a further, distinct criterion: information quantity alone is insufficient
to assess it (Prakash et al., 2020).

Our initial aim was to reproduce selected early evolutionary simulations.
Within the search described in Supplementary Section S1, we did not identify
a single executable package sufficient to reproduce those calculations exactly.
This finding does not extend to all related literature: Berke et al. (2022),
for example, published code and data subsequently used by Mattes (2026).
In the present study, we investigate our own finite model rather than reproduce
the early simulations.

We ask a narrower question: \textbf{when does a representation that retains less
information about the world allow a given task to be solved more accurately?}
Separately, we examine how a charge for a chosen resource changes net
performance—the payoff after subtracting costs.

The model is limited to four equiprobable world states and three binary tasks.
In each task, one answer corresponds to two states and the other to the remaining
two. Within these bounds, we completely enumerate deterministic representations
and action rules—mappings without random choice.

We make two kinds of comparison. In the first, one component of an agent's
construction changes while the others remain fixed. For example, the grouping
of world states can change while the response rule is retained. This comparison
shows the result of a particular replacement under specified constraints.
In the second, we determine the best accuracy available to each representation
by choosing its response rule separately. Here, we compare capabilities after
optimization.

The counting rule also matters. One representation may belong to several
constructions with different sensory filters or different amounts of spare
capacity. We therefore compare two censuses: one includes all such realizations,
and the other counts each representation once.

We then examine how the advantage within a fixed pair depends on the task or
task mixture being evaluated, and on the resource charge under each of four
specified tariffs. All censuses retain cases favoring the less informative
or the more informative member, equal information, and equal payoffs.

The contribution is the explicit, exactly computed set of comparisons and its
reproducible implementation. It brings the effects of action constraints, counting
rules, task mixtures, and four cost assumptions into one inspectable finite model.
The underlying distinctions between information and decision value are established
in earlier theory, including the Blackwell comparison of signals (Blackwell, 1953;
Rauh et al., 2017), and information-cost trade-offs (Marzen \& DeDeo, 2017).
We use these principles to explain the census, not claim them as new discoveries.

\FloatBarrier
\section*{2. Model and methods}

\subsection*{2.1 World, tasks, and payoff}

Let the ordered world be

\[
W=(00,01,10,11), \qquad P(w)=1/4,
\]

with binary action set \(\mathcal{A}=\{0,1\}\). The three balanced task partitions are

\[
y_{first}=(0,0,1,1),\quad
y_{second}=(0,1,0,1),\quad
y_{xor}=(0,1,1,0).
\]

The first task asks for the left bit, the second for the right bit, and exclusive OR
(XOR) asks whether the bits differ. The first two have archived identifiers
\nolinkurl{bit_0} and \nolinkurl{bit_1}. These are all balanced binary task partitions up to
output complement: there are \({4\choose2}=6\) ways to assign label 1 to two states,
and exchanging labels 0 and 1 leaves three partitions.

A correct action scores 1 and an incorrect action scores 0:

\[
u_t(w,a)=\mathbf{1}[a=y_t(w)].
\]

\emph{Payoff} is the average score across the four world states: the proportion of
correct answers. For example, an agent that answers correctly in three of the four
states has a payoff of \(3/4=0.75\), corresponding to 75\% correct answers.
A rule with full access to the world can
score 1 on each task; a constant action scores 1/2.

Figure 1 follows the calculation from world state to representation, action, and
payoff. A separately specified cost may then be subtracted.

\begin{figure}[htbp]
\centering
\includegraphics[width=\linewidth,height=0.76\textheight,keepaspectratio]{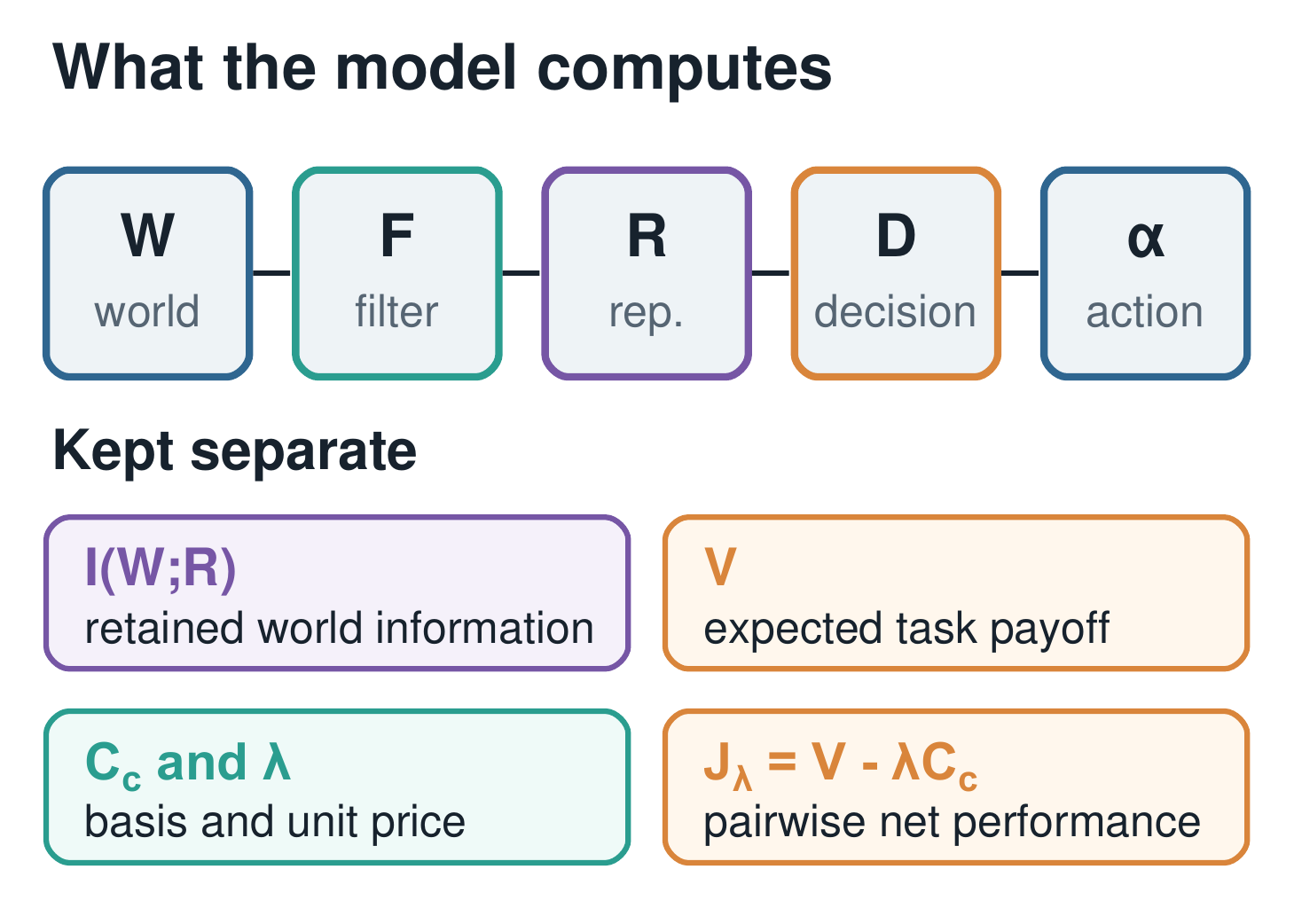}
\par\smallskip
{\small\raggedright \textbf{Figure 1.} The quantities used in the model: retained information, task payoff,
mechanism cost, and net performance after cost.\par}
\end{figure}

\subsection*{2.2 Representations and action rules}

A representation \(R=r(W)\) groups world states into categories. The rule \(r\)
is deterministic: the same state always belongs to the same category.
We denote the number of categories used by \(K_R\).

To give identical groupings identical encodings, we read the states in the specified
order. The first category encountered receives label 0, the next new category
receives label 1, and so on. The labels used form the set
\(Z_r=\{0,\ldots,K_R-1\}\).

For example, \nolinkurl{0011} means that the first two states belong to one category and the
last two to another. This encoding is called a \emph{restricted-growth string} (RGS).
Renaming categories does not create a new representation.

The four world states can be partitioned into categories in 15 ways: one uses one
category, seven use two, six use three, and one uses four.

A \emph{finer} representation preserves every distinction made by a coarser one and may
subdivide its categories further. If at least one category is actually split, the
refinement is \emph{strict}. Such representations are \emph{nested}. If neither of two
representations preserves all the distinctions made by the other, they are
\emph{non-nested}.

An action rule, or \emph{policy}, assigns each category the answer 0 or 1. We denote it
by \(\pi:Z_r\rightarrow\{0,1\}\). A representation with \(K_R\) categories has
\(2^{K_R}\) such rules, and we enumerate them all.

Across all 15 representations, this gives 94 representation-policy combinations.
Each combination is evaluated on three tasks, yielding \(94\times3=282\) result
rows.

The payoff of a given policy and the best payoff available to a representation
are defined as follows:

\[
V_t(r,\pi)=\frac14\sum_{w\in W}
\mathbf{1}[\pi(r(w))=y_t(w)],
\qquad
V_t^*(r)=\max_{\pi}V_t(r,\pi).
\]

The \emph{payoff ceiling} \(V_t^*(r)\) is the highest average score attainable on task
\(t\) with the given representation by choosing the best policy. If several
policies achieve this maximum score, we retain them all.

Mutual information \(I(W;R)\) measures how much knowing the representation category
reduces uncertainty about the world state on average (Shannon, 1948).
We call this quantity retained world information.

Uncertainty is measured by entropy \(H\), expressed in bits. For four equally
likely states, \(H(W)=\log_2 4=2\) bits. Because the world state uniquely determines
the representation category, retained world information equals the entropy of
the representation itself:

\[
I(W;R)=H(R).
\]

However, information about the world does not necessarily help solve a particular
task. We therefore also measure \(I(Y_t;R)\): how much information about the correct
answer \(Y_t\) is contained in the representation.

Finally, the policy maps the representation to an action \(A_{out}\).
The quantity \(I(Y_t;A_{out})\) measures how much information about the correct answer
is contained in the selected action. We thus distinguish information available in
the representation from information reflected in the agent's response.

To test how a policy performs on another task, we first find all policies that are
optimal for the source task. We then evaluate each policy, without changing it, on
the other two tasks. For each source–target task pair, we retain the lowest and
highest payoffs among the source-optimal policies. This shows what the previous
response rules achieve without adjustment.

For 15 representations, three source tasks, and two possible target tasks, this
gives \(15\times3\times2=90\) result rows.

Figure 2 brings together all 15 representations. It shows which states they group
together, how much information they retain, and what accuracy they can achieve.
Here the policy is optimized separately for each task.

\begin{figure}[htbp]
\centering
\includegraphics[width=\linewidth,height=0.76\textheight,keepaspectratio]{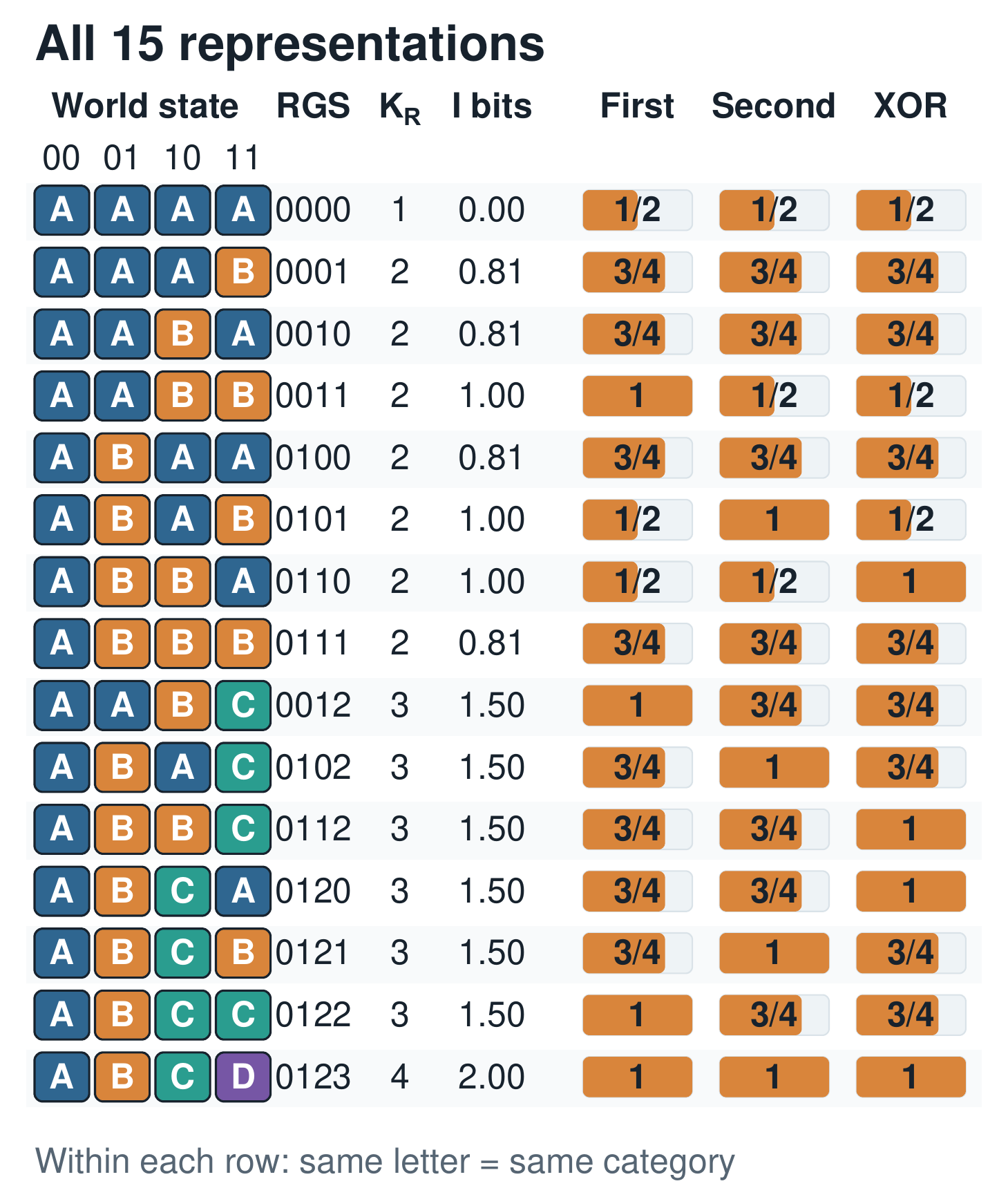}
\par\smallskip
{\small\raggedright \textbf{Figure 2.} All 15 ways to partition the four world states into categories.
Within a row, equal letters and fills indicate states in the same category.
RGS is the grouping code; \(K_R\) is the number of categories used;
\(I\) is retained world information in bits. The orange bars show the best accuracy
on each task after choosing an optimal policy. For example, representation \nolinkurl{0011}
allows the first, left-hand bit of the state to be identified without error.\par}
\end{figure}

\subsection*{2.3 Modules and coded constructions}

Each agent construction specifies which world states the agent can distinguish,
which distinctions it retains in its representation, and how it chooses an action.

The \emph{sensory filter} \(F\) determines which distinctions between world states are
available to the agent. States grouped together by the filter reach the agent as
the same input.

The \emph{representation} \(R\) determines which of these available distinctions are
retained. It may further merge inputs into shared categories, but cannot recover
distinctions unavailable through the filter. For example, if the filter cannot
distinguish states \nolinkurl{00} and \nolinkurl{01}, the representation must also assign them to
the same category.

\emph{Provisioned capacity} \(K\in\{1,2,3,4\}\) limits the number of representation
categories. Not all of this capacity needs to be used: with \(K=4\), for example,
a representation may contain only two categories. We therefore distinguish
provisioned capacity \(K\) from the realized category count \(K_R=|R|\).
Every admissible construction satisfies \(K_R\le K\).

It remains to specify which action the agent chooses for each category.
The \emph{decision code} \(D\) assigns it an answer of 0 or 1. The \emph{action code}
\(\alpha\) either leaves this answer unchanged or reverses it:
0 becomes 1, and 1 becomes 0.

The complete construction thus has five components:

\[
g=(F,K,R,D,\alpha).
\]

For world state \(w\), the representation determines a category, the decision code
assigns it an answer, and the action code turns that answer into the final action:

\[
A_g(w)=\alpha\!\left(D(R(w))\right).
\]

We call the combination of filter, capacity, and representation \((F,K,R)\)
a \emph{module}. It does not yet include an action rule. The \emph{module payoff ceiling}
is the best score under a given objective across all admissible combinations
of \(D,\alpha\). We assess how well the module can perform a task; learning
or evolutionary search for a suitable action rule is not modelled here.

With capacity \(K=1\), there are 15 admissible modules; with \(K=2\), 47;
with \(K=3\), 59; and with \(K=4\), 60. In total, this gives 181 modules.

The decision code assigns each representation category one of two answers:
0 or 1. For \(|R|\) categories, this gives \(2^{|R|}\) choices of decision code.
Each can be combined with two action codes, either keeping or reversing the answer.
Summed over all modules, these choices give 1,424 complete coded constructions.
The detailed count is provided in Supplementary Section S3A.

We use \emph{model genotype} as a label for a complete construction with its chosen
decision and action codes. A \emph{model phenotype} comprises the module and the action
the construction assigns to each world state. Different combinations of codes
can produce the same actions. When the module is also the same, these constructions
have different genotypes but the same phenotype. These terms describe the
construction's design and behaviour, without biological assumptions.

Different encodings are counted separately, even when they specify the same
actions. When determining the payoff ceiling, we also retain all code combinations
that achieve the maximum for the given module and objective.

\subsection*{2.4 Single-task and multi-task objectives}

We evaluate each construction on individual tasks and on combinations of tasks.
For multiple tasks, the overall score is the arithmetic mean of the payoffs.
For example, for a pair of tasks, we add the two payoffs and divide by two.
All tasks within the chosen set receive equal weight.

The three tasks form seven nonempty sets: three single tasks, three distinct pairs,
and one set containing all three tasks. Each set defines a \emph{scoring objective} —
a rule for calculating the overall score.

Let \(S\) denote the chosen task set and \(|S|\) the number of tasks in it.
If \(V_t(g)\) is the payoff of construction \(g\) on task \(t\), its overall score is

\[
V_S(g)=\frac{1}{|S|}\sum_{t\in S}V_t(g).
\]

The same construction \(g\), with an unchanged action rule, is evaluated on every
task in the set. The term \emph{mixture} refers to averaging scores across a fixed set
of tasks, not to a sequence of environmental changes.
Mixtures use exactly the answer labelings in Section 2.1. Independently inverting
one task's answers is not included as an additional mixture variant: with a shared
action rule, it can change the result. Completeness up to answer inversion refers
to the partitions of individual tasks.

\subsection*{2.5 Which pairs are compared?}

Constructions can be compared in different ways. The first question is: what
changes when only one component is replaced and the rest remain fixed? The second
is: which of two modules can achieve a higher payoff when the best action rule is
chosen separately for each? We consider both cases.

In the \textbf{one-component comparison}, two constructions differ in exactly one of
the components \(F,K,R,D,\alpha\). For example, two representations \(R\) can be
compared with the same \(F,K,D,\alpha\). Both constructions must satisfy the
model's constraints.

Replacing only \(R\) preserves the number of categories: the unchanged decision
code \(D\) contains one answer per category. Retained information can still differ.
For example, two categories can divide the four states into groups of sizes
\(3+1\) or \(2+2\). These comparisons change the grouping at a fixed category
count, rather than removing categories from a representation.

For each fixed setting of the other four components, we compare every pair of
admissible alternatives. Order within a pair does not matter: comparing A with B
and B with A counts as one comparison. A replacement need not correspond to a
single mutation; mutation distances and transition rates are not specified here.

For the information/payoff census, only pairs that replace the representation
\(R\) are used from this group.

In the \textbf{all-module-pairs comparison}, we compare any two distinct modules from
the complete set. For each module and each objective, the decision and action
codes that maximize payoff are chosen separately. These maxima are then compared.
This census shows differences in the modules' capabilities after optimization,
rather than the effect of replacing one component.

We also \textbf{check the effect of counting a representation more than once}.
The same representation can belong to several modules with different filters
\(F\) or provisioned capacities \(K\). Some representations therefore appear
more often than others in the all-module-pairs comparison.

In this counting control, modules with the same representation are collapsed
into a single entry. We then compare all pairs of the 15 distinct representations.
The control uses the same saved results and changes only how they are counted,
without adding constructions. It is termed a \emph{quotient census} in the supplement.

It is important to distinguish a pair from its evaluations. A \emph{comparison identity}
is one pair of constructions or modules. A \emph{comparison row} evaluates that pair
under one objective. Each pair is evaluated under all seven objectives, so there
are seven times as many rows as pairs.

\begingroup\small
\renewcommand{\arraystretch}{1.25}
\begin{longtable}{|>{\raggedright\arraybackslash}p{\dimexpr0.440000\linewidth-2\tabcolsep-1.5\arrayrulewidth\relax}|>{\raggedright\arraybackslash}p{\dimexpr0.250000\linewidth-2\tabcolsep-1.5\arrayrulewidth\relax}|>{\raggedright\arraybackslash}p{\dimexpr0.310000\linewidth-2\tabcolsep-1.5\arrayrulewidth\relax}|}
\hline
Census & Number of pairs & Number of rows \\ \hline\endfirsthead
\hline Census & Number of pairs & Number of rows \\ \hline\endhead
One-component replacements, all five types & 9,568 & 66,976 \\ \hline
Of these: representation \(R\) replacements only & 1,416 & 9,912 \\ \hline
All pairs of the 181 modules & 16,290 & 114,030 \\ \hline
All pairs of the 15 representations, without repetitions & 105 & 735 \\ \hline
\end{longtable}
\endgroup

The number of pairs among \(n\) distinct objects is \(n(n-1)/2\):
each pair is counted once.

Figure 3 shows the two comparison methods and the counting control. Each census
has its own set of pairs and total number of comparisons, so the resulting
fractions are reported separately.

\begin{figure}[htbp]
\centering
\includegraphics[width=\linewidth,height=0.76\textheight,keepaspectratio]{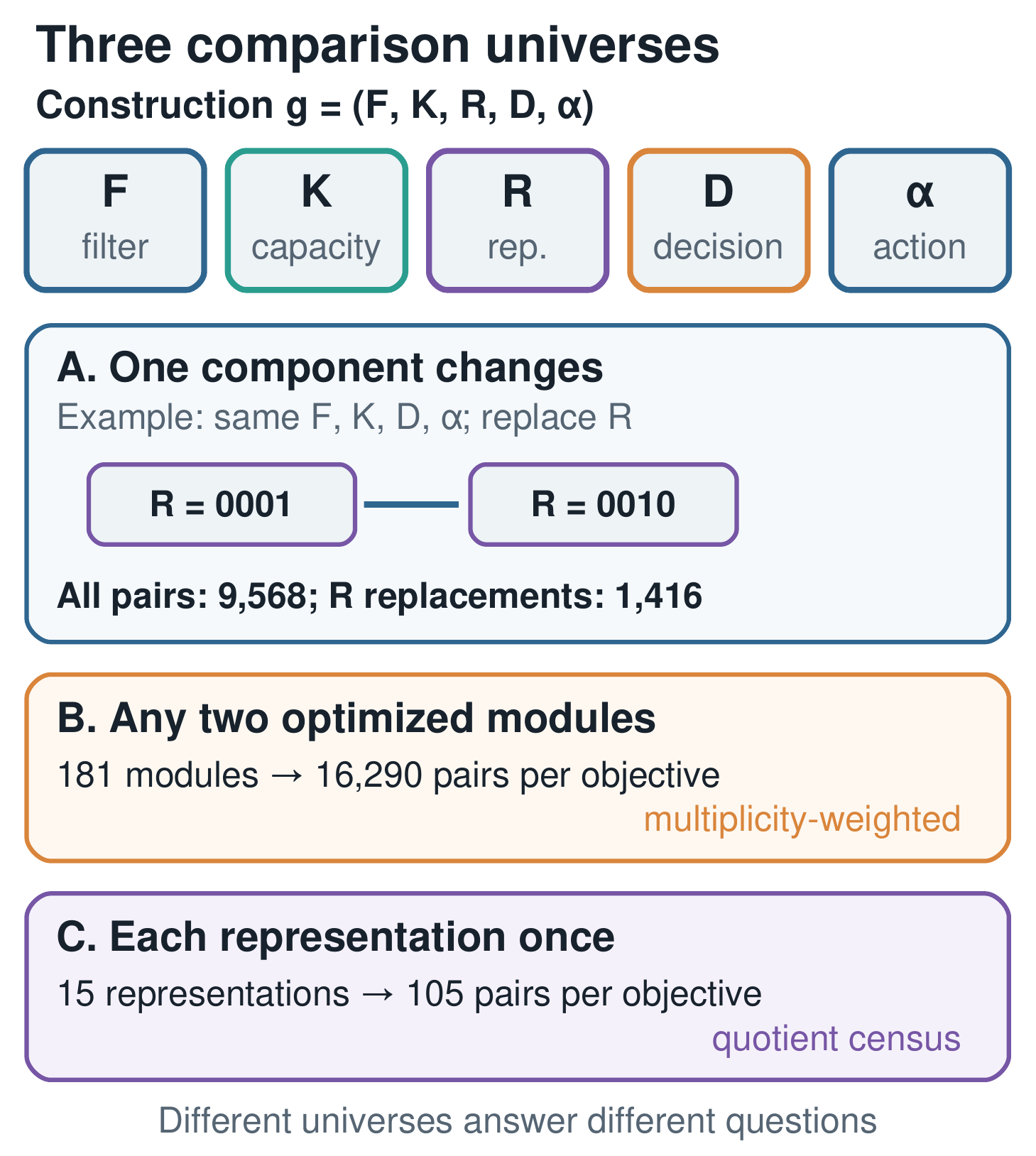}
\par\smallskip
{\small\raggedright \textbf{Figure 3.} Part A compares constructions that differ in one component.
Part B compares the best attainable payoffs of all module pairs; one
representation may belong to several modules. Part C shows the counting control,
in which these repeated realizations are collapsed and pairs of distinct
representations are compared.\par}
\end{figure}

Together, these comparisons show under which conditions lower or higher retained
world information is associated with higher payoff: for a specific representation
replacement or after choosing the best action rule for each module. The
representation-once control allows us to determine how much the fractions of such
cases depend on some representations belonging to more modules than others.

\subsection*{2.6 Information and payoff orders}

We now compare two properties of the members of each pair: how much world
information they retain and what payoff they receive. We ask which member has
the higher payoff — the one retaining less or more information.

We compare information amounts using exact values of \(I(W;R)\). For each
objective, we assign the comparison result to one of four groups:

\begin{enumerate}

\item \textbf{Equal information amounts.} Payoffs may be either equal or different.

\item \textbf{Less information, higher payoff.} The member retaining less world
   information receives a higher payoff than the other.

\item \textbf{More information, higher payoff.} The member retaining more world
   information receives a higher payoff than the other.

\item \textbf{Different information amounts, equal payoffs.} The difference in retained
   information is not accompanied by a payoff difference under this objective.

\end{enumerate}

Each result row belongs to exactly one of these four groups. Equal-information
comparisons are kept separate: neither member can be identified as more or less
informative in these comparisons.

Equal information amounts do not imply identical representations. Values of
\(I(W;R)\) can coincide even when the representations retain different
distinctions between world states.

\subsection*{2.7 Changes of order between objectives}

We next ask whether one member of a pair can receive a higher payoff on an
individual task, while the other receives a higher payoff on a task mixture.

To compare the results, we assign the members the labels "left" and "right".
These labels remain fixed across all objectives. The payoff difference is

\[
\Delta V=V_{left}-V_{right}.
\]

If \(\Delta V>0\), the left member has the higher payoff; if \(\Delta V<0\),
the right member does. At \(\Delta V=0\), the payoffs are equal.

We record a single-task-versus-multi-task reversal if one member strictly
outperforms the other on at least one single task, and the other member strictly
outperforms it on at least one mixture. In other words, the payoff differences
must have opposite nonzero signs. Equal payoffs alone do not count as a reversal.

For module pairs, we also run a broader check: does each member have a strict
advantage under at least one of the seven objectives? Here the opposing results
may concern, for example, two single tasks, rather than necessarily a single task
and a mixture.

\subsection*{2.8 Four separate cost tariffs}

So far, we have compared payoffs without costs. We now ask a different question:
does a construction retain its advantage when a particular resource has a price?

We separately examine four assumptions about which resource carries a cost:

\begingroup\small
\renewcommand{\arraystretch}{1.25}
\begin{longtable}{|>{\raggedright\arraybackslash}p{\dimexpr0.440000\linewidth-2\tabcolsep-1.5\arrayrulewidth\relax}|>{\raggedright\arraybackslash}p{\dimexpr0.250000\linewidth-2\tabcolsep-1.5\arrayrulewidth\relax}|>{\raggedright\arraybackslash}p{\dimexpr0.310000\linewidth-2\tabcolsep-1.5\arrayrulewidth\relax}|}
\hline
Resource priced & Notation & What the quantity measures \\ \hline\endfirsthead
\hline Resource priced & Notation & What the quantity measures \\ \hline\endhead
Sensory access & \(C_F=\lvert F\rvert-1\) & The number of distinguishable inputs beyond one \\ \hline
Provisioned capacity & \(C_K=K-1\) & The number of provisioned categories beyond one, even if not all are used \\ \hline
Representation categories & \(C_R=\lvert R\rvert-1\) & The number of categories actually used beyond one \\ \hline
Retained world information & \(C_I=I(W;R)\) & The amount of retained world information in bits \\ \hline
\end{longtable}
\endgroup

The first three quantities are measured relative to a one-category baseline.
These are hypothetical resource measures, not measurements of energy or metabolism.
Keeping the cost explicit follows the information–cost distinction relevant to
lossy compression (Marzen \& DeDeo, 2017).

We specify the resource amount and its price separately. Let \(x\) denote a
construction or module, \(e\) an objective, and \(c\) the chosen tariff.
The quantity \(C_c(x)\) gives the resource amount being counted. Its hypothetical
price per unit is \(\lambda\ge0\).

The product \(\lambda C_c(x)\) is the cost expressed in payoff units.
Subtracting it from payoff \(V_e(x)\) gives \textbf{net performance}:

\[
J_\lambda(x,e,c)=V_e(x)-\lambda C_c(x).
\]

A high payoff therefore does not guarantee high net performance: costs may
outweigh the advantage.

For a pair, let \(\Delta C\) denote the difference in resource amounts and
\(\Delta J\) the difference in net performance. Both differences use the same
"left minus right" order as \(\Delta V\). Then

\[
\Delta J=\Delta V-\lambda\Delta C.
\]

There are five cases:

\begin{enumerate}

\item \textbf{Equal payoffs and resource amounts.} Net performances are equal at every
   price.

\item \textbf{Different payoffs with equal resource amounts.} The higher-payoff member
   retains its advantage.

\item \textbf{Equal payoffs with different resource amounts.} At zero price, net
   performance is equal; at a positive price, it is higher for the cheaper member.

\item \textbf{The higher-payoff member is also more expensive.} The leader depends on
   the resource price.

\item \textbf{The higher-payoff member is also cheaper.} It retains its advantage at
   every nonnegative price.

\end{enumerate}

For \(\Delta C\ne0\), setting \(\Delta J=0\) gives

\[
\lambda^*=\frac{\Delta V}{\Delta C}.
\]

Only the fourth case has a positive threshold at which the leader changes.
Below \(\lambda^*\), the member with the higher original payoff retains the
advantage. At the threshold, net performances are equal; above it, the cheaper
member has the advantage.

We retain all five cases. Each tariff is evaluated separately: the four costs
are not added together into a single penalty.

\subsection*{2.9 Computation and validation of results}

The finite sets under study, scoring objectives, and cost rules were fixed
in internal configurations before their respective calculations.

Later, during internal manuscript review, we added the representation-once and
nested-refinement controls. We also separately counted pairs whose payoffs differ
under at least one objective. Worked examples were selected by a fixed ordering,
independently of effect size. Algebraically equivalent threshold expressions were
standardized. These additional analyses use the same underlying finite sets;
Supplementary Section S1 describes their chronology and purpose.

Calculations used Python 3.12 and exact rational arithmetic. Information values
were stored as expressions \(q_0+q_3\log_2 3\), with coefficients \(q_0\) and
\(q_3\) represented as exact fractions. This representation allows equality and
order to be determined without rounding.

Tables and figures are based on computed results saved before the publication
summaries were prepared. Summary measures were derived from these records using
unambiguous counting rules.

Validation used separate programs with their own implementations of the
calculations. They reconstructed the full results: all 15 representations and
related policy and transfer sets; all 44,775 modular-analysis rows; and all 916,265
objective- and cost-analysis rows. The checks included threshold signs and all
alternatives attaining the same maximum payoff.

Results matched in every checked row. Supplementary Section S10 provides details
of the validation.

Supplementary tables are numbered in their order of appearance in the Supplement.
Tables S1–S2 document provenance and analysis chronology; S3 gives the worked
information example; S4–S7 give output counts and their derivation; S8–S12 give
comparison denominators and censuses; S13–S15 cover reversals and worked examples;
S16–S17 summarize the separate cost tariffs. Section and table numbers are distinct.

\FloatBarrier
\section*{3. Results}

\subsection*{3.1 Accuracy depends on more than the number of categories}

Consider the task of identifying the first, left-hand bit of the world state.

Representation \nolinkurl{0011} groups the states into two categories: \(\{00,01\}\)
and \(\{10,11\}\). In the first category, the left-hand bit is always 0;
in the second, it is always 1. Assigning the corresponding answer to each category
is therefore enough to solve the task perfectly.

Representation \nolinkurl{0102} has more categories — three: \(\{00,10\}\), \(\{01\}\),
and \(\{11\}\). However, its first category groups states with different
left-hand bits. A single answer cannot be correct for both. Even the best policy
answers correctly in only three of the four states.

Thus, two categories allow a payoff of 1, while three categories with a different
grouping allow only \(3/4\). The reason is visible in the groupings themselves:
they retain different distinctions between world states.

\begingroup\small
\renewcommand{\arraystretch}{1.25}
\begin{longtable}{|>{\raggedright\arraybackslash}p{\dimexpr0.166667\linewidth-2\tabcolsep-1.5\arrayrulewidth\relax}|>{\raggedright\arraybackslash}p{\dimexpr0.166667\linewidth-2\tabcolsep-1.5\arrayrulewidth\relax}|>{\raggedright\arraybackslash}p{\dimexpr0.166667\linewidth-2\tabcolsep-1.5\arrayrulewidth\relax}|>{\raggedright\arraybackslash}p{\dimexpr0.166667\linewidth-2\tabcolsep-1.5\arrayrulewidth\relax}|>{\raggedright\arraybackslash}p{\dimexpr0.166667\linewidth-2\tabcolsep-1.5\arrayrulewidth\relax}|>{\raggedright\arraybackslash}p{\dimexpr0.166667\linewidth-2\tabcolsep-1.5\arrayrulewidth\relax}|}
\hline
Representation & Categories \(K_R\) & World information \(I(W;R)\), bits & Task information \(I(Y;R)\), bits & Information in the action \(I(Y;A_{out})\), bits & Best task payoff \\ \hline\endfirsthead
\hline Representation & Categories \(K_R\) & World information \(I(W;R)\), bits & Task information \(I(Y;R)\), bits & Information in the action \(I(Y;A_{out})\), bits & Best task payoff \\ \hline\endhead
\nolinkurl{0011} & 2 & 1.000 & 1.000 & 1.000 & 1 \\ \hline
\nolinkurl{0102} & 3 & 1.500 & 0.500 & 0.311 & 3/4 \\ \hline
\end{longtable}
\endgroup

The information measures show the same mismatch. Representation \nolinkurl{0102} retains more
information about the world overall, but less about the correct answer than \nolinkurl{0011}.
In an optimal action for \nolinkurl{0102}, information about the correct answer amounts to
approximately 0.311 bits. This value is rounded; exact expressions and both optimal
policies for \nolinkurl{0102} are provided in Supplementary Table S3.

Figure 2 shows that each of the three tasks can be solved perfectly with two
categories if the grouping matches the task. The complete census of comparisons
across all 15 representations shows how common the different cases are within
this finite model. Here, one comparison is one pair of representations evaluated
on one of the three single tasks.

\begin{itemize}

\item In \textbf{12 of 207 comparisons with different category counts}, the representation
  with fewer categories solves the task perfectly, while the one with more
  categories makes errors.

\item In \textbf{66 of 108 comparisons with equal category counts}, the best attainable
  payoffs differ.

\item In \textbf{57 of 243 comparisons with different amounts of world information},
  the representation retaining more world information contains the same amount
  of information about the correct task answer, or less.

\end{itemize}

The example in the table is the first of the 12 cases above in the fixed ordering
of grouping codes. Tasks are considered in the order: first bit, second bit, XOR.

Finally, having the required information and choosing an appropriate action are
distinct conditions for solving a task successfully. We examined how a policy
optimal for the source task performs on another task without adjustment.

Even for a representation that distinguishes all four world states, this unchanged
policy falls short of the best target-task rule in all six directed transfers:
from each of the three tasks to the other two. The representation still contains
all the necessary information, but the policy answers the source-task question.
Solving the new task perfectly requires only choosing the appropriate action rule.

\subsection*{3.2 Refining a representation does not lower its payoff ceiling}

In the preceding example, representations \nolinkurl{0011} and \nolinkurl{0102} group the world
states differently. Each preserves distinctions that the other loses. Such
representations are called \emph{non-nested}.

Now consider a different change: preserve all existing distinctions and add new
ones. We can do this by splitting existing categories into smaller ones, without
merging states from different original categories. The resulting representation
is \emph{finer}. If at least one category is split, the refinement is \emph{strict}.

For example, take representation \nolinkurl{0011} and split category \(\{10,11\}\) into
\(\{10\}\) and \(\{11\}\), leaving category \(\{00,01\}\) unchanged.
For the left-bit task, both new categories can be assigned the same answer, 1.
An additional distinction is available, but the policy need not use it.

In the same way, a finer representation can reproduce the behaviour of any
policy available to the coarser one: simply assign the original answer to every
part of a split category. When the best policy is chosen and cost is not
included, the finer representation's payoff ceiling therefore cannot be lower.
This argument connects to the classical comparison-of-experiments framework,
which compares information sources through the decisions they make available
(Blackwell, 1953).

The complete census of strict-refinement pairs across the three single
tasks contains 135 comparisons. The finer representation has a higher payoff
ceiling in 90 cases and an equal ceiling in 45. There are no cases in which the coarser
representation has a higher ceiling (Supplementary Table S12).

Thus, simply removing distinctions from a given representation does not
increase its best attainable payoff. Strict lower-information/higher-ceiling
cases compare non-nested representations that preserve different distinctions
between world states.

This conclusion concerns payoff \(V^*\) before costs are subtracted. If the
finer representation is more expensive under the chosen cost tariff, its
payoff can be at least as high while its net performance \(J_\lambda\) is
lower after cost is subtracted.

\subsection*{3.3 How coding rules affect the census}

Different encodings can specify identical actions. This happens when every
answer in the decision code \(D\) is complemented and the action code \(\alpha\)
is switched at the same time: the two changes cancel each other out.
For example, an answer of 0 first becomes 1, but the switched action code
turns it back into 0. The final action remains unchanged for every world state.

The 1,424 complete coded constructions therefore implement 712 distinct
combinations of module and action rule across all world states. Each such
combination has exactly two encodings.

Each of the 181 modules is evaluated under seven objectives, giving
\(181\times7=1{,}267\) combinations. In each, we retain all encodings that attain
the highest payoff. For each combination, their number is 2, 4, 8, or 16. It is always
even because every encoding has a partner that specifies the same actions.
The absence of a unique optimal encoding therefore does not necessarily mean
that there are several distinct optimal ways to act.

A different feature of the coding explains the equal counts of the two types
of advantage under representation replacement. The complete census contains
9,912 rows, each evaluating one pair of constructions under one objective:

\begin{itemize}

\item in 5,880 rows, the amount of retained world information is equal;

\item in 792 rows, the lower-information member of the pair has a higher payoff;

\item in another 792 rows, the higher-information member has a higher payoff;

\item in 2,448 rows, information differs but payoffs are equal.

\end{itemize}

As explained in Section 2.5, both members of each pair have the same number of
categories. Their membership and sizes, and hence retained information, can differ.

Why does each of the two opposing advantages occur exactly 792 times?
Take any comparison pair and switch \textbf{only the action code \(\alpha\) in both
constructions}, leaving all other components unchanged. Every action now
becomes its opposite. The representations and the amount of retained
information remain unchanged, but correct answers become incorrect and
incorrect answers become correct.

Each construction's payoff therefore changes from \(V\) to \(1-V\). This
holds for both a single task and a task mixture, which averages the task
payoffs. The payoff difference changes sign:

\[
(1-V_1)-(1-V_2)=-(V_1-V_2).
\]

In other words, if the lower-information member previously had the higher
payoff, the higher-information member has the higher payoff after the switch,
and vice versa.

The full enumeration includes both action codes. Every comparison with
unequal information and unequal payoffs therefore has a partner with the
opposite advantage. There are 1,584 such comparisons: 792 in each direction.
The equality also holds when each objective is counted separately.

The equal counts thus follow from the model's rules: binary actions,
correct/incorrect scoring, and the inclusion of both action codes in the
enumeration. With a different action set, different coding rules, or a
different scoring system, this equality is no longer guaranteed.

\subsection*{3.4 The shares depend on whether we count modules or representations}

We now compare what modules can achieve after choosing the best action rule
for each. Under each objective, we optimize their decision and action codes
separately. In the preceding representation-replacement comparison, these
codes remained fixed.

For this census, it matters whether we count modules or representations.
Two modules can group world states identically while differing in their
sensory filter or provisioned capacity. For example, one module may have
spare capacity for unused categories while another has none. These are
different modules, even though their retained representation is the same.

In the set of modules studied, a representation appears in between 1 and 60 modules.
Some representations therefore enter the all-module comparison repeatedly.
To determine how these repetitions affect the result, we compare two
counting rules:

\begin{itemize}

\item \textbf{All module pairs.} The 181 modules give 16,290 pairs. Each is evaluated
  under seven objectives, yielding 114,030 rows.

\item \textbf{Representation pairs without repeated implementations.} We retain one
  instance of each of the 15 representations. This gives 105 pairs and
  \(105\times7=735\) rows.

\end{itemize}

The second census uses the same saved results. Only the counting rule
changes: differences in sensory filters and provisioned capacity no longer
produce repeated comparisons of the same representations.

The table summarizes the two complete censuses across all seven objectives.
Each data row belongs to one of four classes (Supplementary Tables S10 and S11).

\begingroup\small
\renewcommand{\arraystretch}{1.25}
\begin{longtable}{|>{\raggedright\arraybackslash}p{\dimexpr0.440000\linewidth-2\tabcolsep-1.5\arrayrulewidth\relax}|>{\raggedright\arraybackslash}p{\dimexpr0.250000\linewidth-2\tabcolsep-1.5\arrayrulewidth\relax}|>{\raggedright\arraybackslash}p{\dimexpr0.310000\linewidth-2\tabcolsep-1.5\arrayrulewidth\relax}|}
\hline
Information and payoff-ceiling relationship & All module pairs: rows & Distinct representation pairs: rows \\ \hline\endfirsthead
\hline Information and payoff-ceiling relationship & All module pairs: rows & Distinct representation pairs: rows \\ \hline\endhead
Equal information, regardless of payoff & 31,122 & 168 \\ \hline
Lower information, higher payoff ceiling & 7,344 & 60 \\ \hline
Higher information, higher payoff ceiling & 51,891 & 296 \\ \hline
Unequal information, equal payoff ceilings & 23,673 & 211 \\ \hline
\textbf{Total} & \textbf{114,030} & \textbf{735} \\ \hline
\end{longtable}
\endgroup

Both censuses contain cases in which the lower-information member has the
higher payoff ceiling and cases in which the higher-information member has
the advantage. Under both counting rules, the latter cases outnumber the former.

The shares of these cases nevertheless differ. The first census characterizes
all admissible modules, including different implementations of the same
representation. The second characterizes the ways of grouping world states,
counting each once. Any statement about which advantage occurs more often
must therefore specify the counting rule used to obtain the shares.

Figure 4 compares three censuses: representation replacements, all module
pairs, and representation pairs without repetitions. Results are summed over
all seven objectives. Bar length shows the class's share of all rows in that
census, and the adjacent label gives the exact row count.

\begin{figure}[htbp]
\centering
\includegraphics[width=\linewidth,height=0.76\textheight,keepaspectratio]{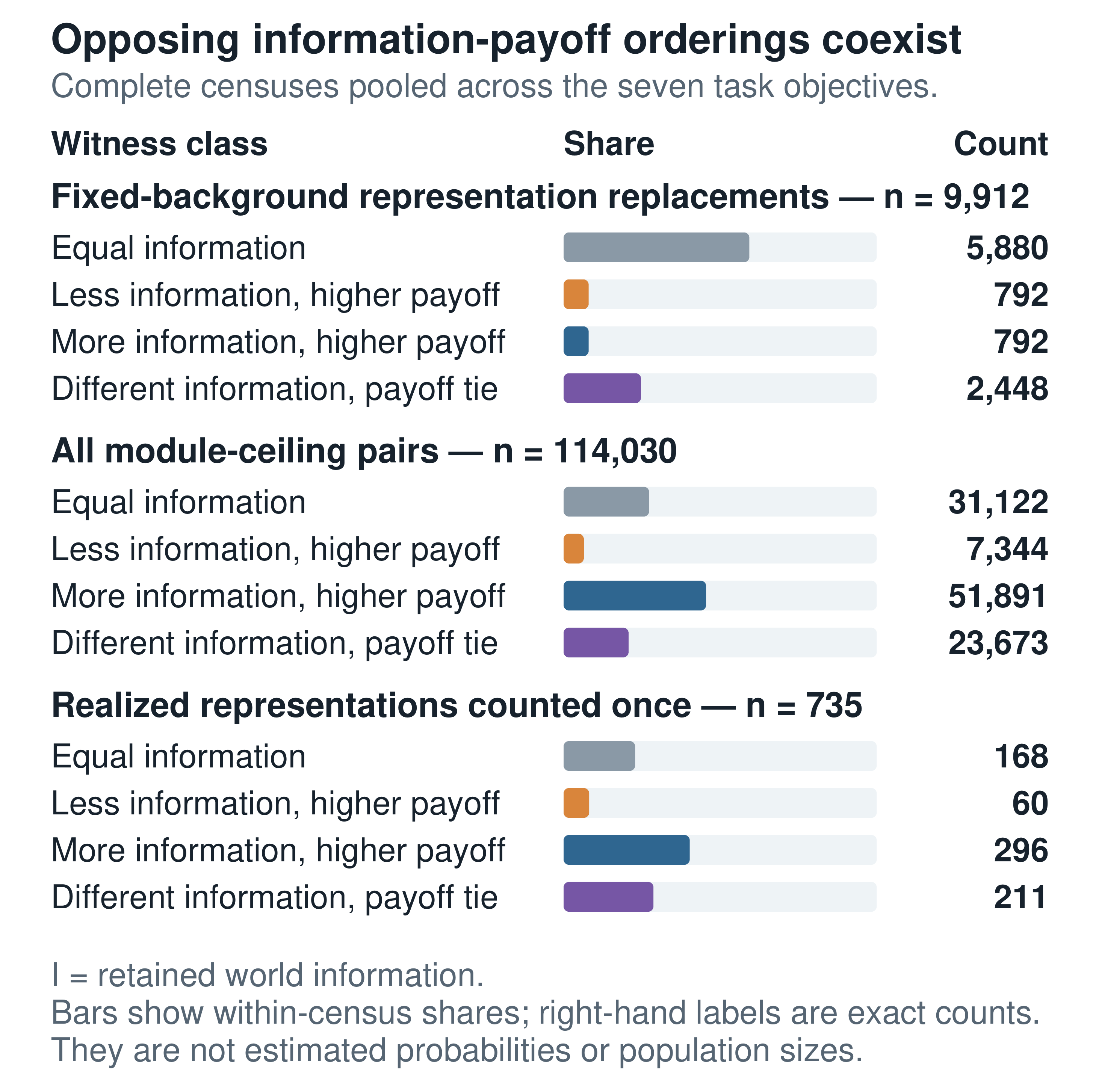}
\par\smallskip
{\small\raggedright \textbf{Figure 4.} Relationships between retained world information and payoff
under three counting rules. For representation replacements, equal counts
of opposing advantages follow from coding symmetry. The all-module
comparison counts different implementations of a representation; the
representation comparison without repetitions counts each representation
once. All four classes are shown separately for each counting rule, including
equal information and equal payoffs at unequal information.\par}
\end{figure}

\subsection*{3.5 The advantage depends on which tasks are evaluated}

The same construction can outperform another on one task and fall behind when
several tasks are evaluated together. The example in Supplementary Table S15
illustrates this. It was selected by a fixed ordering, not by effect size.

The two constructions in this example differ only in their representation.
The left construction uses grouping \nolinkurl{0001}: it combines states \(00,01,10\)
and distinguishes \(11\) separately. The right construction uses grouping
\nolinkurl{0010}: it combines \(00,01,11\) and distinguishes \(10\) separately.
All other components are identical.\footnote{Shared components: \(F=0012\), \(K=2\), \(D=01\);
the action code \(\alpha\) flips the answer. The components are defined
in Section 2.3.} The labels
“left” and “right” remain attached to these constructions under every
objective, as described in Section 2.7.

\begingroup\small
\renewcommand{\arraystretch}{1.25}
\begin{longtable}{|>{\raggedright\arraybackslash}p{\dimexpr0.440000\linewidth-2\tabcolsep-1.5\arrayrulewidth\relax}|>{\raggedright\arraybackslash}p{\dimexpr0.250000\linewidth-2\tabcolsep-1.5\arrayrulewidth\relax}|>{\raggedright\arraybackslash}p{\dimexpr0.310000\linewidth-2\tabcolsep-1.5\arrayrulewidth\relax}|}
\hline
Tasks evaluated & Left construction's payoff & Right construction's payoff \\ \hline\endfirsthead
\hline Tasks evaluated & Left construction's payoff & Right construction's payoff \\ \hline\endhead
Right-bit task & 1/4 & 3/4 \\ \hline
Equal mixture of the left-bit and XOR tasks & 1/2 & 1/4 \\ \hline
\end{longtable}
\endgroup

The right construction has the higher payoff on the single task, whereas
the left has the higher payoff on the mixture. The left-minus-right payoff
difference is \(-1/2\) and \(+1/4\), respectively. Both constructions and
their action rules remain unchanged; only the set of tasks used to evaluate
performance differs.

We count an advantage reversal when at least one single task and at least
one task mixture favour different members of the pair. The payoffs must
differ in both cases: equal payoffs alone do not establish a reversal.

The complete census of 9,568 pairs differing in one component identifies
\textbf{2,116 pairs with an advantage reversal}.

Not all admissible component replacements affect payoff, however. Many pairs
have equal payoffs under all seven objectives. Retaining only pairs whose
payoffs differ under at least one objective leaves 3,592 pairs. These are
called \emph{payoff-active} in the supplementary materials.

The same 2,116 pairs reverse within this set of 3,592. The two denominators
therefore answer different questions:

\begin{itemize}

\item \textbf{2,116 of 9,568:} how many pairs reverse among all admissible
  one-component replacements?

\item \textbf{2,116 of 3,592:} how many reverse among replacements whose payoffs
  differ under at least one objective?

\end{itemize}

Supplementary Table S14 in Section S6 gives the detailed census of active and
remaining pairs for each component type.

For modules, we choose the best action rule separately under each objective.
Among 16,290 module pairs, \textbf{3,672 pairs} reverse their advantage between
a single task and a mixture.

An additional, broader check considers any two of the seven objectives:
does one favour one module while another favours the other? The comparison
can involve two single tasks or two mixtures as well. This condition holds
for \textbf{4,395 module pairs}, including the 3,672 pairs identified above.

\subsection*{3.6 The reversal threshold depends on the chosen cost tariff}

We now keep the task fixed and account for costs. A module may solve the task
more accurately but require more of the resource being priced. Its payoff
advantage must then be weighed against the additional cost.

To do this, we subtract the cost of the resource used from payoff \(V\):

\[
J_\lambda=V-\lambda C.
\]

Here, \(C\) is the resource amount under the chosen tariff, and \(\lambda\)
is the hypothetical price per unit of that resource, expressed in payoff
units. The quantity \(J_\lambda\) is net performance after cost is subtracted.

If the higher-payoff module is also more expensive, it retains its advantage
at a low price. At a sufficiently high price, the extra payoff no longer
compensates for the additional cost, and the cheaper module becomes the
better option. The boundary is an exact positive threshold,

\[
\lambda^*=\frac{\Delta V}{\Delta C}>0.
\]

At this price, net performances are equal. Below the threshold, the
higher-payoff module has the advantage; above it, the lower-cost module does.

For each of the four tariffs, we separately examine the complete finite set
of 114,030 module-pair/objective rows. The table reports how many rows have
such a positive threshold.

\begingroup\small
\renewcommand{\arraystretch}{1.25}
\begin{longtable}{|>{\raggedright\arraybackslash}p{\dimexpr0.440000\linewidth-2\tabcolsep-1.5\arrayrulewidth\relax}|>{\raggedright\arraybackslash}p{\dimexpr0.250000\linewidth-2\tabcolsep-1.5\arrayrulewidth\relax}|>{\raggedright\arraybackslash}p{\dimexpr0.310000\linewidth-2\tabcolsep-1.5\arrayrulewidth\relax}|}
\hline
Resource priced & Resource amount \(C\) & Rows with a positive threshold \\ \hline\endfirsthead
\hline Resource priced & Resource amount \(C\) & Rows with a positive threshold \\ \hline\endhead
Sensory distinctions & \(C_F=\lvert F\rvert-1\) & 31,415 \\ \hline
Provisioned capacity & \(C_K=K-1\) & 30,414 \\ \hline
Representation categories & \(C_R=\lvert R\rvert-1\) & 45,951 \\ \hline
Retained world information & \(C_I=I(W;R)\) & 51,891 \\ \hline
\end{longtable}
\endgroup

The four tariffs are separate alternatives; their costs are not added
together. Both the counts and the threshold values differ. Exact ranges and
summary statistics are reported in Supplementary Table S17 in Section S7.

Consider an example for the left-bit task. This is the first pair with a
positive threshold under the information tariff in the fixed module
enumeration order.

The left module uses representation \nolinkurl{0000}, grouping all four world states
into one category. It retains 0 bits of information and attains payoff
\(1/2\).

The right module uses representation \nolinkurl{0001}, distinguishing state \(11\)
from the other three. It attains a higher payoff, \(3/4\), and retains

\[
I(W;R)=2-\tfrac34\log_2 3
\]

bits of information.\footnote{Left module: \(F=0000,\ K=1,\ R=0000\);
right module: \(F=0001,\ K=2,\ R=0001\).}

The right module's payoff advantage is \(1/4\). The cost of its additional
information exactly offsets this advantage at a price of

\[
\lambda^*
=\frac{1/4}{\,2-\tfrac34\log_2 3\,}
=\frac{1}{8-3\log_2 3}
\approx 0.3082
\]

payoff units per bit. Above this price, the left module has higher net
performance. Its payoff still equals \(1/2\): it has not become more accurate
at the task, but it is the better option after costs are included.

The count of 51,891 under the information tariff matches the number of
higher-information/higher-ceiling cases in Section 3.4. These are the same
comparison rows. With \(C_I=I(W;R)\), more information automatically means
more of the resource being priced. Each such case therefore has a positive
threshold at which the advantage reverses. The equal counts follow from
the tariff's definition and are not a separate, independent confirmation
of the result.

Figure 5 brings together the two ways of changing the advantage. In part A,
the set of evaluated tasks differs. In part B, the task remains fixed,
but the price of the chosen resource is included. The first comparison
concerns payoffs; the second concerns net performance after cost.

\begin{figure}[htbp]
\centering
\includegraphics[width=\linewidth,height=0.76\textheight,keepaspectratio]{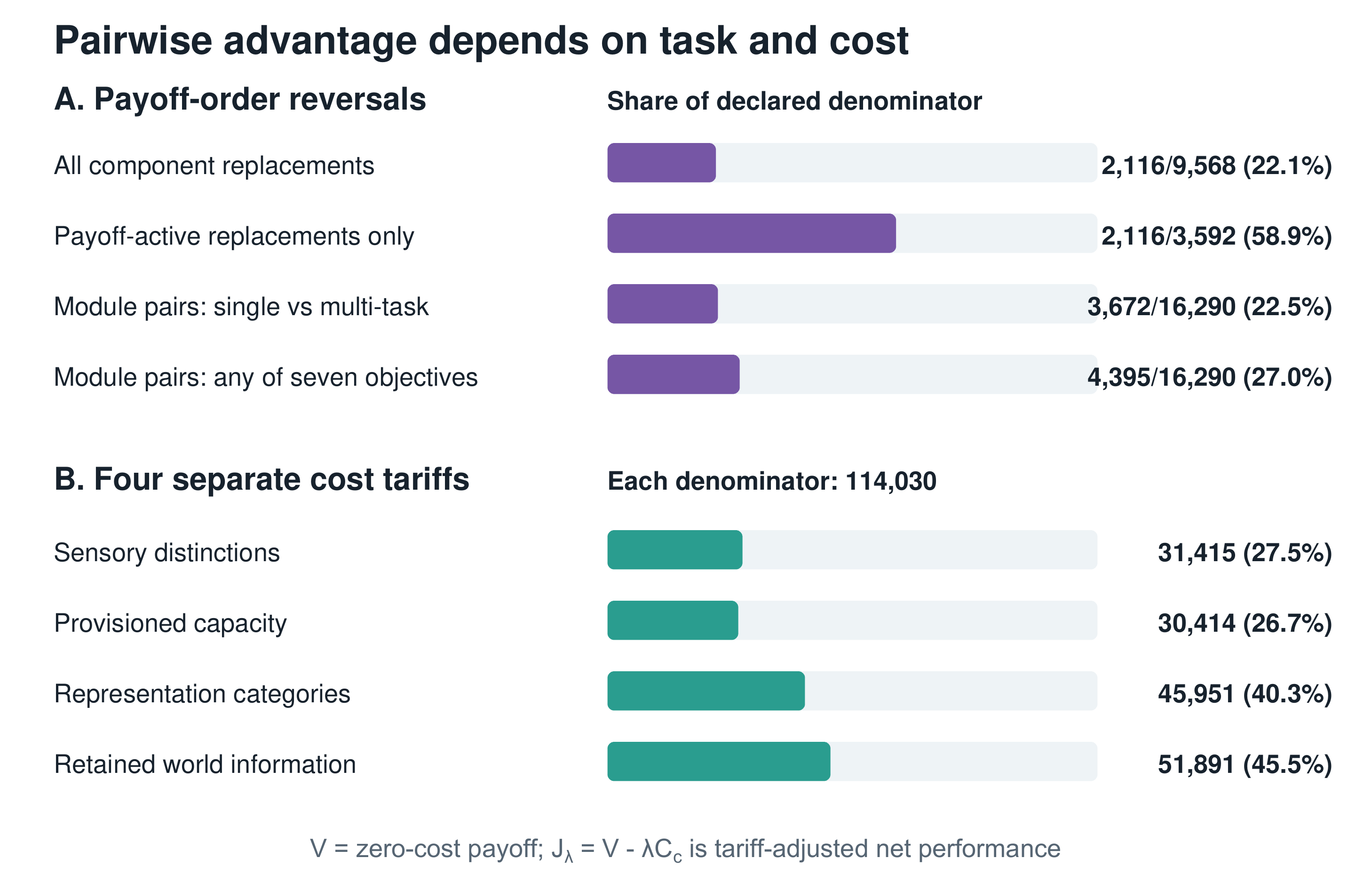}
\par\smallskip
{\small\raggedright \textbf{Figure 5.} Part A shows the number of pairs whose advantage reverses
between a single task and a mixture. For one-component replacements, counts
are reported among all admissible pairs and among payoff-active pairs;
for modules, the broader check across any two objectives is also shown.
Part B shows the number of comparisons with a positive reversal threshold
under each of the four separate tariffs.\par}
\end{figure}

\FloatBarrier
\section*{4. Discussion}

\subsection*{4.1 Task performance depends on which distinctions a representation preserves}

The example in Section 3.1 shows why more retained world information does not
always allow greater task accuracy. A representation can preserve many
distinctions while merging precisely the states that require different answers.
Another representation may retain less information overall but distinguish
states that matter for the task, thereby providing a higher payoff ceiling.

A different situation arises when a representation preserves all previous
distinctions and only adds new ones. A policy can leave the additional
distinctions unused and reproduce the previous actions. With no cost and the
best policy chosen, the payoff ceiling therefore does not decrease.
The lower-information representation's advantage in the first example comes
from retaining different distinctions, not simply from deleting some
information from a given representation.

Accounting for costs introduces another basis for a lower-information
advantage. If the lower-information module is cheaper under the chosen tariff,
the savings may outweigh its task-performance deficit. Its payoff \(V\) can
then be lower while its net performance after cost, \(J_\lambda\), is higher.
The advantage here comes from lower costs, not greater accuracy.

How often the different cases occur also depends on which comparisons are
included in the census. Under representation replacement, lower-information/higher-payoff
and higher-information/higher-payoff cases occur in equal numbers because of
coding symmetry. In module comparisons, the shares depend on how many different
modules implement each representation. Reported shares must therefore specify
which constructions were compared and what counted as a distinct member of
a comparison pair.

The set of evaluated tasks matters as well. The same constructions with
unchanged action rules can have different advantages depending on whether
a single task or a mixture of tasks is evaluated. The usefulness of retained
distinctions depends not only on the representation but also on the tasks
to be solved.

The separation of information quantity from decision value is established in
earlier work (Strouse \& Schwab, 2017; Rauh et al., 2017). In particular, our
nested-refinement result is a deterministic special case of the Blackwell order,
not a new theorem. This work also builds on studies involving multiple objectives
or environmental change (Martínez, 2019; Angelucci et al., 2021; Berke et al., 2022;
Mattes, 2026).
The atlas brings these questions together in one finite model with a complete
enumeration of admissible constructions. This makes it possible to compare
lower- and higher-information advantages, retain cases of equal payoff, and explain which
results follow from the model and counting rules.

\subsection*{4.2 How the results relate to Hoffman's ideas}

Hoffman's ideas led us to ask whether better task performance necessarily
requires more information about the world. Our atlas contains cases in which
a representation retains less information but supports a higher payoff.
In this sense, the results are consistent with successful action based on
selectively retained information.

The atlas also contains cases in which more informative representations have
the advantage. For nested representations, a more definite result holds:
preserving all previous distinctions and adding new ones cannot lower the
best attainable payoff before costs are included. The results are therefore
compatible with both the loss of some information and the preservation of
useful distinctions. The answer depends on which distinctions are retained,
which tasks are evaluated, and which costs are included.

To relate these findings to the fitness-beats-truth argument, it is important
to distinguish the quantities involved. In this study, payoff is a score for
solving a task. It does not measure survival or reproduction. To determine
how an observed advantage affects the spread of constructions in a population,
we would need to specify a link between payoff and reproduction and a model
of evolutionary dynamics.

Retained world information \(I(W;R)\) also has a specific meaning: it measures
how much knowing the representation reduces uncertainty about the state of
the modelled world. The number of bits alone does not determine whether a
description of the world is true or identify which relations it preserves.
Two representations with the same amount of information can distinguish
different states and help solve different tasks.

Assessing structure preservation requires additional criteria. For example,
one could separately test whether a representation reflects causal
relations—what affects what—or spatial relations—what is adjacent to what.
This kind of question, considered by Prakash et al. (2020), differs from our
comparison of information amount and payoff.

Finally, the fitness-beats-truth theorem is formulated for particular
strategy classes and mathematical conditions (Prakash et al., 2021).
Our atlas uses a different finite model, so its results should not be
treated as a direct test of that theorem. It supplies a worked finite comparison
in which the effects of action rules, counting, and costs can be inspected before
introducing a population model.

\subsection*{4.3 What the finite atlas contributes}

The atlas is intended as a reference for checking comparisons, rather than as a
new theory of compression or a new optimization method. For example, a reader can
check whether a reported lower-information advantage persists after the action
rule is optimized, or whether a census changes when repeated realizations are
counted only once. The saved results make these checks possible within the stated
model; their usefulness as a benchmark for other methods remains to be evaluated.

This focus differs from mapping the best compression-relevance trade-offs
(Tegmark \& Wu, 2020; Tan et al., 2022). Our pairwise cost thresholds include
comparisons between constructions that need not be globally optimal. Beating one
competitor at a given price does not establish superiority over every admissible
alternative. The four tariffs expose different modelling choices, not newly
discovered mechanisms of resource use.

The model's small size makes it possible to examine all admissible constructions
and all comparisons specified by its rules. This reveals the cases in which
a less informative representation has the advantage, those in which a more
informative one does, and those in which the results are equal.

Exact arithmetic reliably distinguishes equality from differences in the
results. Equality in this census is established from exact values: a match
between rounded numbers is not enough. This matters both for comparing payoffs and for identifying
the price at which net performances become equal.

Complete enumeration connects the numerical results to the model's rules.
Supplementary Section S8 gives assumptions and short proofs for four properties:
the ability to ignore additional distinctions, action-inversion symmetry,
the effect of repeated encodings on pair counts, and the emergence of a
threshold when information is priced. These properties help explain the
relationships found in the atlas.

The effect of repeated encodings deserves a separate explanation. Even multiplying the number of
encodings of every model phenotype by the same factor can change the
comparison shares. Here, a phenotype is a module together with its actual
actions in every world state. If the census includes pairs of equivalent
encodings of one phenotype, it counts two kinds of pairs: within a phenotype
and between different phenotypes. Duplication increases their counts in
different proportions. The same increase in encoding multiplicity therefore
need not preserve the shares of the results.

These mathematical explanations help account for the observed relationships.
The resulting shares describe the complete census of the specified finite
set of comparisons under the chosen counting rules.

Treating costs separately also matters. The number of distinguishable
sensory inputs, provisioned capacity, the number of used categories, and
retained information describe different resources. The four tariffs allow
the consequences of each cost assumption to be examined separately.

Applying these tariffs to an organism or device would require identifying
the physical process responsible for the costs, their measurement units,
and their relationship to the chosen model quantity. The hypothetical price
could then be linked to measurements of a specific system.

Related questions arise in learning visual representations. For example,
Franca trains nested portions of a feature vector so that its truncated
versions also remain useful (Venkataramanan et al., 2026). Our atlas offers
a simple formal model in which retained distinctions, available decision
rules, and representation cost can be examined separately. Extending these
arguments to learned systems requires accounting for the training procedure
and the limitations of the decision rule used.

\subsection*{4.4 From comparing constructions to population evolution}

The atlas allows us to compare how well different constructions perform tasks.
The next step for the five-component constructions in this paper is to test how
these differences affect the spread of variants in a reproducing population.

This requires specifying how payoff relates to reproductive fitness—expected
reproductive success. Population size and the rules of inheritance, mutation,
and reproduction must also be defined. It would then be possible to estimate
the probability of fixation, when one variant occupies the entire population,
or the shares of different variants over the long term.

For example, fitness could be assigned positive values such that higher payoff
always corresponds to higher fitness. This would preserve pairwise order:
the construction with higher payoff would also have higher fitness. That
relationship alone, however, would not determine its fixation probability
or long-run share of the population.

The atlas results help identify further questions. Would different encodings
with identical behaviour be inherited as distinct variants? Which mutations
would connect them to other constructions? How should genetic drift—random
changes in variant frequencies during reproduction—be modelled?
If tasks change over time, the sequence of those changes must also be specified:
a task mixture with fixed weights does not describe such a sequence.

The exact comparisons in the atlas can serve as benchmarks for this model.
The new test would show when an advantage in task performance leads to the
spread of a construction and when population dynamics change the outcome.

\FloatBarrier
\section*{5. Limitations}

\textbf{Which model was completely enumerated.}
Complete enumeration applies to a model with four equiprobable world states
and three binary tasks. The tasks are balanced: each of the two answers is
correct in two of the four states. Representations and policies are
deterministic—the same input always produces the same output.

The results do not cover larger worlds, unequal state probabilities,
random noise in representations, learning, or interaction with the environment
through a body and its movements. Task mixtures have fixed weights: they do
not describe environmental change over time, memory, or adaptation.

\textbf{What the quantities measure.}
Retained world information measures how much a representation reduces
uncertainty about the world state. Payoff is average answer accuracy,
and net performance is payoff minus the specified cost.

These quantities do not provide a general assessment of a representation's
veridicality. Nor does the model establish how payoff relates to reproductive
success or how cost relates to actual energy use, memory use, or computation
time.

\textbf{How the rules affect comparisons and counting.}
When one component is replaced, the other four remain fixed. The result
therefore characterizes that replacement in a particular construction;
it does not establish a universal effect of the component.

The enumeration rule specifies which constructions are compared, but not
how often mutation could turn one into the other or how many mutational
steps this would require.

In the module census, a representation may be counted repeatedly; in the
representation-once census, it is counted only once. The resulting shares
depend on that choice. They describe a complete census of the specified
comparisons, not the prevalence of such constructions among natural agents.

\textbf{Which evolutionary conclusions remain beyond the model.}
A pairwise payoff advantage does not show whether a rare variant will spread,
occupy the entire population, disappear, or coexist with other variants.
Such conclusions require a population model of the kind outlined in Section 4.4.

This work continues a research question from the fitness–truth literature,
but does not reproduce published evolutionary simulations.

\FloatBarrier
\section*{6. Reproducibility and AI use}

An accompanying package supports checking and repeating the calculations.
It contains the manuscript, supplement, source tables, vector figures,
calculation code, separate programs for checking the results, and selected tests.
The package also includes model specifications, fixed calculation settings,
and records of the exact software dependency versions.

The archive preserves the original computational results and the summaries
derived from them. Null results, equal payoffs or net performances, and order
reversals are included alongside cases where one construction has a strict
advantage over another.

The package includes a manifest listing the files, their locations, sizes,
and checksums. A checksum serves as a digital fingerprint of a file's contents
and allows changes to be detected. Supplementary Sections S10–S13 describe
the checks, the links between claims and results, and the workflow.

The package was validated in the Python (CPython) environment on Windows
described in the manifest. Producing byte-identical files on other operating
systems has not yet been independently verified. The texts of primary
research papers are not included in the package.

OpenAI Codex was used for source organization, code development and testing,
manuscript drafting and editing, and internal consistency checks.
The first author reviewed the scientific decisions and interpretations
and is responsible for the manuscript's content. The use of AI in internal
checks is described in more detail in the supplement.

\FloatBarrier
\section*{7. Conclusion}

To assess how well a representation supports a task, knowing only how much
world information it retains is not enough. The particular distinctions
it preserves matter. In the four-state model studied here, a representation
retaining less information can achieve higher accuracy if it better preserves
the distinctions needed for the task. Cases favouring more information
and cases of equal accuracy also occur.

Simply discarding distinctions, however, does not improve the best attainable
result when cost is excluded. If one representation preserves all the
distinctions of another and adds new ones, an optimal policy can ignore
the additions and repeat the earlier actions. Additional distinctions
therefore do not reduce the payoff ceiling.

Advantage also depends on the task and costs. Changing the task mixture
can change which of two unchanged constructions receives the higher payoff.
Once cost is included, the net-performance advantage can shift to a cheaper
construction even though its answer accuracy remains unchanged. The shares
of such cases depend on which pairs are compared and whether repeated
implementations of one representation are counted separately.

The relationship between information and payoff must therefore be considered
together with the retained distinctions, task, available decision rules,
cost, and counting method. These general distinctions are established in earlier
theory. Our contribution is an exact, reproducible account of their consequences
in one fully enumerated model, including equal outcomes, coding symmetries, and
alternative counting rules. The saved comparisons can serve as reference cases
for subsequent modelling; they do not establish a new general advantage of
information loss or an outcome of population evolution.

\FloatBarrier
\section*{Author contributions}

Aleksandr M. Shtraukh-Adamsky led the study and made the principal
contributions to conceptualization, model development, computational experiment
design, computational implementation, validation and analysis of results, and
manuscript drafting and editing. The use of AI tools is described in Section 6.

Vladimir M. Adamsky, Irina G. Voloshina, and Artemiy D. Spitsyn
contributed to the discussion and development of the research concept,
models, and computational experiment designs.

All authors have read and approved the manuscript.

\FloatBarrier
\section*{Short glossary}

\textbf{Representation and category.} A representation groups world states. States in
the same category receive the same internal label, so the representation does not
distinguish them.

\textbf{Retained world information.} \(I(W;R)\) measures how much knowing the category
reduces uncertainty about the world state on average. It is measured in bits.

\textbf{Bit of information.} A unit of information. Learning which of two equiprobable
possibilities has occurred provides one bit. Average information can be fractional;
the introduction explains the value of 1.5 bits with a worked example.

\textbf{Task and objective.} A task specifies which answer is correct. An objective
evaluates accuracy on one task or average accuracy across several tasks with
specified weights.

\textbf{Policy.} A rule assigning an action to each representation category.

\textbf{Payoff and payoff ceiling.} \(V\) is average accuracy on a specified task or task
mixture. For example, \(V=3/4\) corresponds to 75\% correct answers. Equal payoffs
mean equal accuracy under the same evaluation conditions.
\(V^*\) is the highest accuracy attainable by choosing the best allowed
policy for the given representation or module.

\textbf{Module.} A combination of three components: a sensory filter defining the
distinctions available at the input, provisioned capacity, and a representation.
Decision and action codes have not yet been selected.

\textbf{Capacity and category count.} \(K\) is the maximum allowed number of
representation categories. \(K_R\) is the number actually used. For example,
a representation with capacity \(K=4\) may use only two categories.

\textbf{Nested representations.} Two representations are nested if one preserves every
distinction made by the other. The finer representation may further split the
coarser one's categories, but does not merge states that the coarser one already
distinguishes.

\textbf{Comparison pair and row.} A pair consists of two constructions or two modules
being compared. A comparison row evaluates that pair under one objective. One
pair evaluated under seven objectives therefore produces seven rows.

\textbf{Complete census.} A count of every case specified by the chosen rules, such as
all admissible constructions or all pairs being compared.

\textbf{Resource, price, and net performance.} \(C_c\) is the resource quantity under the
chosen tariff, and \(\lambda\) is its unit price. The cost \(\lambda C_c\) is
subtracted from payoff to obtain net performance:

\[
J_\lambda=V-\lambda C_c.
\]

\textbf{Crossing threshold.} The price \(\lambda^*\) at which the two compared endpoints
have equal net performance. If the more accurate endpoint is also more expensive,
it has the advantage below this threshold; above it, the cheaper endpoint does.

An accompanying reader guide provides a longer glossary and answers to common
questions.

\FloatBarrier
\section*{References}

Angelucci, A., Fano, V., Ferretti, G., Macrelli, R., \& Tarozzi, G. (2021).
Evolutionary dynamics and accurate perception: Critical realism as an empirically
testable hypothesis. \emph{Philosophia Scientiæ, 25-2}, 157–178.
\url{https://doi.org/10.4000/philosophiascientiae.2960}

Berke, M. D., Walter-Terrill, R., Jara-Ettinger, J., \& Scholl, B. J. (2022).
Flexible goals require that inflexible perceptual systems produce veridical
representations: Implications for realism as revealed by evolutionary simulations.
\emph{Cognitive Science, 46}(10), e13195. https://doi.org/10.1111/cogs.13195

Blackwell, D. (1953). Equivalent comparisons of experiments. \emph{The Annals of
Mathematical Statistics, 24}(2), 265–272. https://doi.org/10.1214/aoms/1177729032

Hoffman, D. D. (2019). \emph{The case against reality: Why evolution hid the truth from
our eyes}. W. W. Norton \& Company.

Mark, J. T., Marion, B. B., \& Hoffman, D. D. (2010). Natural selection and veridical
perceptions. \emph{Journal of Theoretical Biology, 266}(4), 504–515.
\url{https://doi.org/10.1016/j.jtbi.2010.07.020}

Martínez, M. (2019). Usefulness drives representations to truth: A family of
counterexamples to Hoffman's interface theory of perception. \emph{Grazer Philosophische
Studien, 96}(3), 319–341. https://doi.org/10.1163/18756735-09603004

Marzen, S. E., \& DeDeo, S. (2017). The evolution of lossy compression. \emph{Journal of
the Royal Society Interface, 14}(130), 20170166.
\url{https://doi.org/10.1098/rsif.2017.0166}

Mattes, J. (2026). The case against \emph{The case against reality}. \emph{Biological Theory}.
\url{https://doi.org/10.1007/s13752-026-00531-0}

Prakash, C., Stephens, K. D., Hoffman, D. D., Singh, M., \& Fields, C. (2021).
Fitness beats truth in the evolution of perception. \emph{Acta Biotheoretica, 69},
319–341. https://doi.org/10.1007/s10441-020-09400-0

Prakash, C., Fields, C., Hoffman, D. D., Prentner, R., \& Singh, M. (2020).
Fact, fiction, and fitness. \emph{Entropy, 22}(5), 514.
\url{https://doi.org/10.3390/e22050514}

Rauh, J., Banerjee, P. K., Olbrich, E., Jost, J., Bertschinger, N., \& Wolpert, D.
(2017). Coarse-graining and the Blackwell order. \emph{Entropy, 19}(10), 527.
\url{https://doi.org/10.3390/e19100527}

Shannon, C. E. (1948). A mathematical theory of communication. \emph{Bell System
Technical Journal, 27}(3–4), 379–423, 623–656.
July installment: \url{https://doi.org/10.1002/j.1538-7305.1948.tb01338.x}
October installment: \url{https://doi.org/10.1002/j.1538-7305.1948.tb00917.x}

Strouse, D. J., \& Schwab, D. J. (2017). The deterministic information bottleneck.
\emph{Neural Computation, 29}(6), 1611–1630. https://doi.org/10.1162/NECO\_a\_00961

Tan, A. K., Tegmark, M., \& Chuang, I. L. (2022). Pareto-optimal clustering with
the primal deterministic information bottleneck. \emph{Entropy, 24}(6), 771.
\url{https://doi.org/10.3390/e24060771}

Tegmark, M., \& Wu, T. (2020). Pareto-optimal data compression for binary
classification tasks. \emph{Entropy, 22}(1), 7. https://doi.org/10.3390/e22010007

Venkataramanan, S., Pariza, V., Salehi, M., Knobel, L., Ramzi, E., Gidaris, S.,
Bursuc, A., \& Asano, Y. M. (2026). Franca: Nested Matryoshka clustering for
scalable visual representation learning. \emph{Proceedings of the IEEE/CVF Conference
on Computer Vision and Pattern Recognition}, 10533–10544.
\url{https://openaccess.thecvf.com/content/CVPR2026/papers/Venkataramanan_Franca_Nested_Matryoshka_Clustering_for_Scalable_Visual_Representation_Learning_CVPR_2026_paper.pdf}

\FloatBarrier
\clearpage
\begin{center}
{\Large\bfseries Supplementary Material\par}
\end{center}

\FloatBarrier
\section*{When Does Less Information Pay? An Exact Finite Atlas of Representation, Payoff, and Cost}

\textbf{Status:} Supplement to preprint version 1.0. \textbf{License:} CC BY 4.0.\\

\textbf{Task:} \nolinkurl{TASK-PUB-005}\\

\textbf{Claim:} \nolinkurl{CLM-PUB-001}\\

\textbf{Epistemic status:} \nolinkurl{E}

\FloatBarrier
\section*{S1. Scope and provenance}

The paper synthesizes only accepted Stage 7.2A–7.2C results:

Table S1 documents the scope and fixed inputs of the three analyses.

\textbf{Table S1. Scope and provenance of the three exact analyses.}

\begingroup\small
\renewcommand{\arraystretch}{1.25}
\begin{longtable}{|>{\raggedright\arraybackslash}p{\dimexpr0.070000\linewidth-2\tabcolsep-1.5\arrayrulewidth\relax}|>{\raggedright\arraybackslash}p{\dimexpr0.150000\linewidth-2\tabcolsep-1.5\arrayrulewidth\relax}|>{\raggedright\arraybackslash}p{\dimexpr0.150000\linewidth-2\tabcolsep-1.5\arrayrulewidth\relax}|>{\raggedright\arraybackslash}p{\dimexpr0.220000\linewidth-2\tabcolsep-1.5\arrayrulewidth\relax}|>{\raggedright\arraybackslash}p{\dimexpr0.410000\linewidth-2\tabcolsep-1.5\arrayrulewidth\relax}|}
\hline
Stage & Decision & Assumption & Claim & Exact config SHA-256 \\ \hline\endfirsthead
\hline Stage & Decision & Assumption & Claim & Exact config SHA-256 \\ \hline\endhead
7.2A & \nolinkurl{RDR-0028} & \nolinkurl{ASM-INT-0002} & \nolinkurl{CLM-INT-BOTTLENECK-001} & \nolinkurl{a1feca94512b19023e518a6000de7901cebb17e72d387873cc0d397ad94e6622} \\ \hline
7.2B & \nolinkurl{RDR-0029} & \nolinkurl{ASM-INT-0003} & \nolinkurl{CLM-INT-GENOTYPE-001} & \nolinkurl{0f2fe87d1347fcfd648bcdceb39395df91617ef2ad0926227ca86f6f5dd9298e} \\ \hline
7.2C & \nolinkurl{RDR-0030} & \nolinkurl{ASM-INT-0004} & \nolinkurl{CLM-INT-SELECTION-BRIDGE-001} & \nolinkurl{5d3f2e25a80b60d9e89d4abbfaed5c4c295f44eb9f55f2bb00a3558cef40c8ef} \\ \hline
\end{longtable}
\endgroup

The paper is a project-defined exact finite extension. It is not a direct
reproduction of Mark et al. (2010), Prakash et al. (2021), or any source genetic
algorithm. The Hoffman book is motivation only. The renewed source audit found the
public Berke et al. (2022) OSF code/data package and therefore limits the recorded
reproducibility gap to the older materials examined.

The study was not publicly preregistered. The primary finite sets of constructions
and comparisons, the objectives, and the tariffs were fixed in the exact stage
configurations before their executions. The following saved-evidence analyses
were added later during manuscript review:

Table S2 records the purpose of the later saved-evidence analyses.

\textbf{Table S2. Analyses added during manuscript review.}

\begingroup\small
\renewcommand{\arraystretch}{1.25}
\begin{longtable}{|>{\raggedright\arraybackslash}p{\dimexpr0.440000\linewidth-2\tabcolsep-1.5\arrayrulewidth\relax}|>{\raggedright\arraybackslash}p{\dimexpr0.250000\linewidth-2\tabcolsep-1.5\arrayrulewidth\relax}|>{\raggedright\arraybackslash}p{\dimexpr0.310000\linewidth-2\tabcolsep-1.5\arrayrulewidth\relax}|}
\hline
Later analysis & Reason added & Scientific effect \\ \hline\endfirsthead
\hline Later analysis & Reason added & Scientific effect \\ \hline\endhead
representation-once quotient & expose multiplicity weighting in the 181-module census & alternative counting measure over the same realized representations; no new model rows \\ \hline
strict nested-refinement census & test whether lower-information witnesses arise by merely deleting distinctions & narrows the mechanism claim; no new model rows \\ \hline
active/inert denominator decomposition & distinguish all categorical replacements from those that ever change payoff & clarifies denominators only \\ \hline
information-decomposition and order examples & make the information and reversal mechanisms readable & selected mechanically, never by effect magnitude \\ \hline
exact threshold canonicalization & collapse algebraically equivalent symbolic quotients & changes the number of distinct spellings/classes, not any saved threshold row \\ \hline
\end{longtable}
\endgroup

The dated search procedure behind the limited reproducibility-gap statement is
packaged as
\nolinkurl{provenance/PAPER-001-hoffman-motivation-reproducibility-gap-audit.md}.

\FloatBarrier
\section*{S2. Exact finite model}

World states are \nolinkurl{00, 01, 10, 11}, each with probability \nolinkurl{1/4}. Actions are \nolinkurl{0,1}.
The three task label vectors are:

\begin{Verbatim}[fontsize=\small]
bit_0 = (0,0,1,1)
bit_1 = (0,1,0,1)
xor   = (0,1,1,0)
\end{Verbatim}

In reader-facing prose, \nolinkurl{bit_0} is the \textbf{first-coordinate} task (the left bit in the
declared world ordering) and \nolinkurl{bit_1} is the \textbf{second-coordinate} task (the right
bit). These three tasks exhaust balanced binary labelings of four states up to output
complement: \nolinkurl{choose(4,2)/2 = 3}.

This completeness statement concerns individual task partitions. Multi-task
objectives use the three answer labelings listed above without independent
inversions. Because the same action rule serves every task in a mixture, inverting
one task's answers can change the payoff of a fixed representation.

Payoff is one for a correct action and zero otherwise. A deterministic
representation is a label-free partition of the four states encoded as a
restricted-growth string. The 15 canonical identities are available in
\nolinkurl{source_tables/fig2_representation_atlas.csv}.

For deterministic \nolinkurl{R=r(W)}, retained world information is \nolinkurl{I(W;R)=H(R)} with base-2
entropy. Exact values use \nolinkurl{q0 + q3*log2(3)}, where \nolinkurl{q0} and \nolinkurl{q3} are rational
coefficients stored in each symbolic information certificate. Display decimals are
non-normative.

\subsection*{Exact values for the introductory example}

The following saved example concerns the first-coordinate task. Policy bits
are listed in category-label order. Both optimal policies for \nolinkurl{0102} have
the information-in-action value shown. The example was selected by the
fixed ordering described in main-text Section 3.1.

Table S3 gives the exact values and optimal policies for the introductory example.

\textbf{Table S3. Exact information decomposition of the introductory example.}

\begingroup\small
\renewcommand{\arraystretch}{1.25}
\begin{longtable}{|>{\raggedright\arraybackslash}p{\dimexpr0.440000\linewidth-2\tabcolsep-1.5\arrayrulewidth\relax}|>{\raggedright\arraybackslash}p{\dimexpr0.250000\linewidth-2\tabcolsep-1.5\arrayrulewidth\relax}|>{\raggedright\arraybackslash}p{\dimexpr0.310000\linewidth-2\tabcolsep-1.5\arrayrulewidth\relax}|}
\hline
Quantity & Representation \nolinkurl{0011} & Representation \nolinkurl{0102} \\ \hline\endfirsthead
\hline Quantity & Representation \nolinkurl{0011} & Representation \nolinkurl{0102} \\ \hline\endhead
Used categories & \nolinkurl{2} & \nolinkurl{3} \\ \hline
Category labels in world order & \nolinkurl{[0,0,1,1]} & \nolinkurl{[0,1,0,2]} \\ \hline
Retained world information, bits & \nolinkurl{1} & \nolinkurl{3/2} \\ \hline
Task-relevant information, bits & \nolinkurl{1} & \nolinkurl{1/2} \\ \hline
All optimal policies & \nolinkurl{[[0,1]]} & \nolinkurl{[[0,0,1],[1,0,1]]} \\ \hline
Information in an optimal action, bits & \nolinkurl{1} & \nolinkurl{(3/2)+(-3/4)*log2(3)} \\ \hline
Best task payoff & \nolinkurl{1} & \nolinkurl{3/4} \\ \hline
\end{longtable}
\endgroup

The machine-readable values and selection rule are in
\nolinkurl{source_tables/table_s5_information_decomposition_witness.csv}.

\FloatBarrier
\section*{S3. Complete output counts}

\subsection*{Stage 7.2A}

Table S4 lists the saved output counts for the representation atlas.

\textbf{Table S4. Output counts for the representation atlas.}

\begingroup\small
\renewcommand{\arraystretch}{1.25}
\begin{longtable}{|>{\raggedright\arraybackslash}p{\dimexpr0.680000\linewidth-2\tabcolsep-1.5\arrayrulewidth\relax}|>{\raggedright\arraybackslash}p{\dimexpr0.320000\linewidth-2\tabcolsep-1.5\arrayrulewidth\relax}|}
\hline
Table & Rows \\ \hline\endfirsthead
\hline Table & Rows \\ \hline\endhead
representations & 15 \\ \hline
representation-policy-task & 282 \\ \hline
representation-task ceilings & 45 \\ \hline
frozen transfer & 90 \\ \hline
active-category thresholds & 315 \\ \hline
information-rate thresholds & 315 \\ \hline
failures & 0 \\ \hline
\end{longtable}
\endgroup

Counterexample counts: 12 smaller-\nolinkurl{K_R} perfect/larger-\nolinkurl{K_R} imperfect
representation-pair/task comparisons out of 207 different-\nolinkurl{K_R} comparisons; 66
same-\nolinkurl{K_R} unequal-ceiling comparisons out of 108; 57 of 243 unequal-\nolinkurl{I(W;R)}
representation-pair/task comparisons retain more world information without greater
task-relevant information; six full-state frozen-policy target losses.

\subsection*{Stage 7.2B}

Table S5 lists the saved output counts for the modular-construction atlas.

\textbf{Table S5. Output counts for the modular-construction atlas.}

\begingroup\small
\renewcommand{\arraystretch}{1.25}
\begin{longtable}{|>{\raggedright\arraybackslash}p{\dimexpr0.680000\linewidth-2\tabcolsep-1.5\arrayrulewidth\relax}|>{\raggedright\arraybackslash}p{\dimexpr0.320000\linewidth-2\tabcolsep-1.5\arrayrulewidth\relax}|}
\hline
Table & Rows \\ \hline\endfirsthead
\hline Table & Rows \\ \hline\endhead
compatible modules & 181 \\ \hline
genotypes & 1,424 \\ \hline
genotype-task & 4,272 \\ \hline
module frozen-transfer envelopes & 1,086 \\ \hline
contextual one-locus contrasts & 28,704 \\ \hline
epistasis rectangles & 9,108 \\ \hline
failures & 0 \\ \hline
\end{longtable}
\endgroup

\subsubsection*{S3A. Derivation of the central library counts}

For construction \nolinkurl{g=(F,K,R,D,alpha)}, the realized action is
\nolinkurl{A_g(w)=alpha(D(R(w)))}. Compatibility requires \nolinkurl{R} to coarsen \nolinkurl{F} and
\nolinkurl{|R| <= K}.

Table S6 gives the derivation of the central construction counts.

\textbf{Table S6. Combinatorial derivation of the central construction counts.}

\begingroup\small
\renewcommand{\arraystretch}{1.25}
\begin{longtable}{|>{\raggedright\arraybackslash}p{\dimexpr0.230000\linewidth-2\tabcolsep-1.5\arrayrulewidth\relax}|>{\raggedright\arraybackslash}p{\dimexpr0.770000\linewidth-2\tabcolsep-1.5\arrayrulewidth\relax}|}
\hline
Count & Exact derivation \\ \hline\endfirsthead
\hline Count & Exact derivation \\ \hline\endhead
181 compatible modules & capacity subtotals \nolinkurl{15 + 47 + 59 + 60 = 181} for \nolinkurl{K=1,2,3,4} \\ \hline
1,424 coded constructions & the 181 modules have realized-\nolinkurl{|R|} distribution \nolinkurl{60, 96, 24, 1}; therefore \nolinkurl{60*2^2 + 96*2^3 + 24*2^4 + 1*2^5 = 1,424}, where the extra factor 2 is the action code \\ \hline
9,568 one-coordinate identities & unordered alternative pairs within every fixed four-component background: \nolinkurl{3,264 + 1,320 + 1,416 + 2,856 + 712 = 9,568} for \nolinkurl{F, K, R, D, alpha} \\ \hline
\end{longtable}
\endgroup

These are combinatorial counts in the declared library. They are not sampled
abundances or population sizes.

An epistasis rectangle fixes a background and two binary component substitutions,
then records the non-additive contrast
\nolinkurl{epsilon = V_11 - V_10 - V_01 + V_00}. It is a project-defined contextual
finite-difference diagnostic, not biological epistasis or a population parameter.

The 1,424 genotypes encode 712 distinct module-by-realized-action phenotypes, each
with two neutral decision/action encodings. Complementing the decision code and
flipping the action code preserves the realized action map, so this twofold
multiplicity is a coding invariant.

\subsection*{Stage 7.2C}

Table S7 lists the saved output counts for the objective and cost analyses.

\textbf{Table S7. Output counts for the objective and cost analyses.}

\begingroup\small
\renewcommand{\arraystretch}{1.25}
\begin{longtable}{|>{\raggedright\arraybackslash}p{\dimexpr0.680000\linewidth-2\tabcolsep-1.5\arrayrulewidth\relax}|>{\raggedright\arraybackslash}p{\dimexpr0.320000\linewidth-2\tabcolsep-1.5\arrayrulewidth\relax}|}
\hline
Table & Rows \\ \hline\endfirsthead
\hline Table & Rows \\ \hline\endhead
genotype-environment profiles & 9,968 \\ \hline
module-environment envelopes & 1,267 \\ \hline
registered one-coordinate/environment comparisons & 66,976 \\ \hline
module pair-environment comparisons & 114,030 \\ \hline
registered one-coordinate threshold rows & 267,904 \\ \hline
module threshold rows & 456,120 \\ \hline
total scientific rows & 916,265 \\ \hline
failures & 0 \\ \hline
\end{longtable}
\endgroup

\FloatBarrier
\section*{S4. Environments}

The seven environments are all nonempty equal-weight subsets of the three tasks:

\begin{Verbatim}[fontsize=\small]
stable_bit_0
stable_bit_1
stable_xor
mix_bit_0_bit_1
mix_bit_0_xor
mix_bit_1_xor
mix_all_tasks
\end{Verbatim}

Mixtures are stationary expectations, not temporal switching processes.

\FloatBarrier
\section*{S5. Comparison universes and exact witness counts}

The same comparison pair can contribute one row per objective. The following
inventory gives the denominators used in the article.

Table S8 distinguishes comparison pairs from objective-specific rows.

\textbf{Table S8. Comparison pairs, objectives, and row denominators.}

\begingroup\small
\renewcommand{\arraystretch}{1.25}
\begin{longtable}{|>{\raggedright\arraybackslash}p{\dimexpr0.250000\linewidth-2\tabcolsep-1.5\arrayrulewidth\relax}|>{\raggedright\arraybackslash}p{\dimexpr0.250000\linewidth-2\tabcolsep-1.5\arrayrulewidth\relax}|>{\raggedright\arraybackslash}p{\dimexpr0.250000\linewidth-2\tabcolsep-1.5\arrayrulewidth\relax}|>{\raggedright\arraybackslash}p{\dimexpr0.250000\linewidth-2\tabcolsep-1.5\arrayrulewidth\relax}|}
\hline
Comparison set & Pairs & Objectives & Comparison rows \\ \hline\endfirsthead
\hline Comparison set & Pairs & Objectives & Comparison rows \\ \hline\endhead
All registered one-component replacements & 9,568 & 7 & 66,976 \\ \hline
Representation, decision-code, and action-code family subtotal & 4,984 & 7 & 34,888 \\ \hline
Actually payoff-active registered replacements & 3,592 & 7 & 25,144 \\ \hline
Registered representation-component replacements & 1,416 & 7 & 9,912 \\ \hline
Unrestricted module-ceiling pairs & 16,290 & 7 & 114,030 \\ \hline
Realized representations counted once & 105 & 7 & 735 \\ \hline
\end{longtable}
\endgroup

Full definitions are in \nolinkurl{source_tables/table_s6_denominator_map.csv}.

\subsection*{Registered representation-component replacements}

Across 9,912 rows:

The two representations in each pair have the same number of categories: the
fixed decision code has one bit per used category. Unequal retained information
in this census comes from different block sizes at a fixed category count,
not from comparing a representation with a strict refinement of itself.

Table S9 reports the four classes for representation replacements.

\textbf{Table S9. Information and payoff classes for representation replacements.}

\begingroup\small
\renewcommand{\arraystretch}{1.25}
\begin{longtable}{|>{\raggedright\arraybackslash}p{\dimexpr0.680000\linewidth-2\tabcolsep-1.5\arrayrulewidth\relax}|>{\raggedright\arraybackslash}p{\dimexpr0.320000\linewidth-2\tabcolsep-1.5\arrayrulewidth\relax}|}
\hline
Class & Count \\ \hline\endfirsthead
\hline Class & Count \\ \hline\endhead
equal world information & 5,880 \\ \hline
lower world information, higher payoff & 792 \\ \hline
higher world information, higher payoff & 792 \\ \hline
unequal world information, payoff tie & 2,448 \\ \hline
\end{longtable}
\endgroup

\subsection*{Module-ceiling comparisons}

Across 114,030 rows:

Table S10 reports the four classes for module-ceiling comparisons.

\textbf{Table S10. Information and payoff-ceiling classes for module pairs.}

\begingroup\small
\renewcommand{\arraystretch}{1.25}
\begin{longtable}{|>{\raggedright\arraybackslash}p{\dimexpr0.680000\linewidth-2\tabcolsep-1.5\arrayrulewidth\relax}|>{\raggedright\arraybackslash}p{\dimexpr0.320000\linewidth-2\tabcolsep-1.5\arrayrulewidth\relax}|}
\hline
Class & Count \\ \hline\endfirsthead
\hline Class & Count \\ \hline\endhead
equal world information & 31,122 \\ \hline
lower world information, higher payoff & 7,344 \\ \hline
higher world information, higher payoff & 51,891 \\ \hline
unequal world information, payoff tie & 23,673 \\ \hline
\end{longtable}
\endgroup

These are complete censuses of the specified finite sets of comparisons.
They are not probabilities or empirical frequencies.

\subsection*{Representation-once quotient}

The 181 modules weight realized representations unequally: a realized map has between
1 and 60 compatible sensory-filter/capacity encodings. A deterministic aggregation of
the immutable saved rows therefore counts each realized representation once in each
environment. Across 735 comparisons:

Table S11 reports the census with each representation counted once.

\textbf{Table S11. Information and payoff-ceiling classes with each representation counted once.}

\begingroup\small
\renewcommand{\arraystretch}{1.25}
\begin{longtable}{|>{\raggedright\arraybackslash}p{\dimexpr0.680000\linewidth-2\tabcolsep-1.5\arrayrulewidth\relax}|>{\raggedright\arraybackslash}p{\dimexpr0.320000\linewidth-2\tabcolsep-1.5\arrayrulewidth\relax}|}
\hline
Class & Count \\ \hline\endfirsthead
\hline Class & Count \\ \hline\endhead
equal world information & 168 \\ \hline
lower world information, higher payoff & 60 \\ \hline
higher world information, higher payoff & 296 \\ \hline
unequal world information, payoff tie & 211 \\ \hline
\end{longtable}
\endgroup

This quotient was added during manuscript review to expose the counting measure, not
to replace an unfavourable headline. It is reported beside the original module census
and is derived deterministically from the same immutable rows.

\subsection*{Nested-refinement control}

For a strict refinement, every state pair merged by the finer representation is also
merged by the coarser representation. Across the complete 135
strict-refinement/task comparisons:

Table S12 reports the strict nested-refinement control.

\textbf{Table S12. Strict nested-refinement control.}

\begingroup\small
\renewcommand{\arraystretch}{1.25}
\begin{longtable}{|>{\raggedright\arraybackslash}p{\dimexpr0.680000\linewidth-2\tabcolsep-1.5\arrayrulewidth\relax}|>{\raggedright\arraybackslash}p{\dimexpr0.320000\linewidth-2\tabcolsep-1.5\arrayrulewidth\relax}|}
\hline
Relation of reoptimized ceilings & Count \\ \hline\endfirsthead
\hline Relation of reoptimized ceilings & Count \\ \hline\endhead
finer representation higher & 90 \\ \hline
tie & 45 \\ \hline
coarser representation higher & 0 \\ \hline
\end{longtable}
\endgroup

The zero follows from policy-set inclusion: the finer representation can imitate any
policy available to the coarser one. Strict lower-information/higher-ceiling
witnesses therefore require non-nested partitions in this atlas.

\FloatBarrier
\section*{S6. Reversal observable}

For a fixed left/right identity, let \nolinkurl{delta_value=V_left-V_right}. The registered
single-task-versus-multi-task diagnostic requires one strictly positive sign in the
single-task set and one strictly negative sign in the multi-task set, or
the reverse. Zero establishes neither side. The orientation is never changed by
environment.

Table S13 reports the single-task versus task-mixture reversal counts.

\textbf{Table S13. Single-task versus task-mixture order reversals.}

\begingroup\small
\renewcommand{\arraystretch}{1.25}
\begin{longtable}{|>{\raggedright\arraybackslash}p{\dimexpr0.440000\linewidth-2\tabcolsep-1.5\arrayrulewidth\relax}|>{\raggedright\arraybackslash}p{\dimexpr0.250000\linewidth-2\tabcolsep-1.5\arrayrulewidth\relax}|>{\raggedright\arraybackslash}p{\dimexpr0.310000\linewidth-2\tabcolsep-1.5\arrayrulewidth\relax}|}
\hline
Universe & Unique identities & Single-task-vs-multi-task reversals \\ \hline\endfirsthead
\hline Universe & Unique identities & Single-task-vs-multi-task reversals \\ \hline\endhead
registered one-coordinate comparisons & 9,568 & 2,116 \\ \hline
module pairs & 16,290 & 3,672 \\ \hline
\end{longtable}
\endgroup

The broader “both signs anywhere among seven environments” module diagnostic is
4,395/16,290 and is not interchangeable with 3,672.

The 9,568 denominator includes two component families that are structurally
payoff-inert in the registered composition: 3,264 sensory-filter identities and 1,320
capacity identities. A deterministic decomposition of the immutable rows is:

Table S14 separates payoff-active and inactive comparisons by component.

\textbf{Table S14. Payoff-active and inactive comparisons by component.}

\begingroup\small
\renewcommand{\arraystretch}{1.25}
\begin{longtable}{|>{\raggedright\arraybackslash}p{\dimexpr0.320000\linewidth-2\tabcolsep-1.5\arrayrulewidth\relax}|>{\raggedright\arraybackslash}p{\dimexpr0.130000\linewidth-2\tabcolsep-1.5\arrayrulewidth\relax}|>{\raggedright\arraybackslash}p{\dimexpr0.200000\linewidth-2\tabcolsep-1.5\arrayrulewidth\relax}|>{\raggedright\arraybackslash}p{\dimexpr0.170000\linewidth-2\tabcolsep-1.5\arrayrulewidth\relax}|>{\raggedright\arraybackslash}p{\dimexpr0.180000\linewidth-2\tabcolsep-1.5\arrayrulewidth\relax}|}
\hline
Component family & Identities & Active in ≥1 objective & Inert in all 7 & Reversals \\ \hline\endfirsthead
\hline Component family & Identities & Active in ≥1 objective & Inert in all 7 & Reversals \\ \hline\endhead
sensory filter & 3,264 & 0 & 3,264 & 0 \\ \hline
provisioned capacity & 1,320 & 0 & 1,320 & 0 \\ \hline
representation & 1,416 & 748 & 668 & 676 \\ \hline
decision code & 2,856 & 2,494 & 362 & 1,272 \\ \hline
action code & 712 & 350 & 362 & 168 \\ \hline
representation/decision/action family subtotal & 4,984 & 3,592 & 1,392 & 2,116 \\ \hline
actually payoff-active identities & 3,592 & 3,592 & 0 & 2,116 \\ \hline
\end{longtable}
\endgroup

The registered graph contains every categorical alternative at one component and
fixed background. It does not define a transition process, mutation rate, or distance.

\subsection*{Worked examples}

The first two rows evaluate the same fixed pair under different objectives.
The third row is the separate cost example from main-text Section 3.6.

Table S15 provides the saved objective-reversal and cost-threshold examples.

\textbf{Table S15. Worked objective-reversal and information-cost examples.}

\begingroup\small
\renewcommand{\arraystretch}{1.25}
\begin{longtable}{|>{\raggedright\arraybackslash}p{\dimexpr0.200000\linewidth-2\tabcolsep-1.5\arrayrulewidth\relax}|>{\raggedright\arraybackslash}p{\dimexpr0.200000\linewidth-2\tabcolsep-1.5\arrayrulewidth\relax}|>{\raggedright\arraybackslash}p{\dimexpr0.200000\linewidth-2\tabcolsep-1.5\arrayrulewidth\relax}|>{\raggedright\arraybackslash}p{\dimexpr0.200000\linewidth-2\tabcolsep-1.5\arrayrulewidth\relax}|>{\raggedright\arraybackslash}p{\dimexpr0.200000\linewidth-2\tabcolsep-1.5\arrayrulewidth\relax}|}
\hline
Comparison & Objective & Left payoff & Right payoff & Left minus right \\ \hline\endfirsthead
\hline Comparison & Objective & Left payoff & Right payoff & Left minus right \\ \hline\endhead
Fixed construction pair & Second-coordinate task & 1/4 & 3/4 & -1/2 \\ \hline
Same construction pair & Equal first-coordinate/XOR mixture & 1/2 & 1/4 & 1/4 \\ \hline
Module pair & First-coordinate task; information tariff & 1/2 & 3/4 & -1/4 \\ \hline
\end{longtable}
\endgroup

For the construction pair, the common components are \nolinkurl{F=0012}, \nolinkurl{K=2},
\nolinkurl{D=01}, and the bit-flip action code. The left representation is \nolinkurl{0001};
the right is \nolinkurl{0010}. For the module pair, the left module is
\nolinkurl{F=0000, K=1, R=0000} and the right is \nolinkurl{F=0001, K=2, R=0001}.
Their information costs are respectively \nolinkurl{0} and \nolinkurl{(2)+(-3/4)*log2(3)}.
The positive threshold is \nolinkurl{1/((8)+(-3)*log2(3))}, approximately
\nolinkurl{0.308155726702}. Exact identities, costs, and the unchanged selection
rules are in \nolinkurl{source_tables/table_s7_worked_order_examples.csv}.

\FloatBarrier
\section*{S7. Tariffs and exact threshold algebra}

Four costs are evaluated independently:

\begin{Verbatim}[fontsize=\small]
C_sensory = |F|-1
C_capacity = K-1
C_rep_support = |R|-1
C_rep_information = I(W;R)
\end{Verbatim}

For each tariff, \nolinkurl{V} is zero-cost task payoff, \nolinkurl{C_c} is the amount of the named
mechanism basis, \nolinkurl{lambda} is its hypothetical unit price, and
\nolinkurl{J_lambda=V-lambda*C_c} is tariff-adjusted net performance. \nolinkurl{J_lambda} is not called
payoff. The exact pairwise boundary is
\nolinkurl{lambda*=delta_value/delta_cost} when cost differs. Five exhaustive classes are
retained: coincident; equal-cost payoff dominance; zero-price tie followed by
lower-cost dominance; positive crossing; payoff-and-cost dominance.

Module positive-crossing counts:

Table S16 reports positive crossings and payoff-and-cost dominance separately for each tariff.

\textbf{Table S16. Positive crossings and payoff-and-cost dominance by tariff.}

\begingroup\small
\renewcommand{\arraystretch}{1.25}
\begin{longtable}{|>{\raggedright\arraybackslash}p{\dimexpr0.440000\linewidth-2\tabcolsep-1.5\arrayrulewidth\relax}|>{\raggedright\arraybackslash}p{\dimexpr0.250000\linewidth-2\tabcolsep-1.5\arrayrulewidth\relax}|>{\raggedright\arraybackslash}p{\dimexpr0.310000\linewidth-2\tabcolsep-1.5\arrayrulewidth\relax}|}
\hline
Tariff & Positive crossings & Payoff-and-cost dominance \\ \hline\endfirsthead
\hline Tariff & Positive crossings & Payoff-and-cost dominance \\ \hline\endhead
sensory support & 31,415 & 10,983 \\ \hline
capacity provision & 30,414 & 16,035 \\ \hline
representation support & 45,951 & 1,404 \\ \hline
representation information & 51,891 & 7,344 \\ \hline
\end{longtable}
\endgroup

The tariffs are never pooled and none is interpreted as measured biological cost.

Positive-threshold distributions:

Table S17 summarizes the positive threshold distributions.

\textbf{Table S17. Distributions of positive module-pair thresholds.}

\begingroup\small
\renewcommand{\arraystretch}{1.25}
\begin{longtable}{|>{\raggedright\arraybackslash}p{\dimexpr0.200000\linewidth-2\tabcolsep-1.5\arrayrulewidth\relax}|>{\raggedright\arraybackslash}p{\dimexpr0.200000\linewidth-2\tabcolsep-1.5\arrayrulewidth\relax}|>{\raggedright\arraybackslash}p{\dimexpr0.200000\linewidth-2\tabcolsep-1.5\arrayrulewidth\relax}|>{\raggedright\arraybackslash}p{\dimexpr0.200000\linewidth-2\tabcolsep-1.5\arrayrulewidth\relax}|>{\raggedright\arraybackslash}p{\dimexpr0.200000\linewidth-2\tabcolsep-1.5\arrayrulewidth\relax}|}
\hline
Tariff & Distinct values & Minimum & Median & Maximum \\ \hline\endfirsthead
\hline Tariff & Distinct values & Minimum & Median & Maximum \\ \hline\endhead
sensory support & 8 & 1/36 & 1/4 & 1/2 \\ \hline
capacity provision & 8 & 1/36 & 1/6 & 1/2 \\ \hline
representation support & 6 & 1/24 & 1/4 & 1/2 \\ \hline
representation information & 17 & ≈0.0833 & ≈0.3082 & ≈1.3247 \\ \hline
\end{longtable}
\endgroup

Exact symbolic information thresholds are in
\nolinkurl{source_tables/table_s4_threshold_distributions.csv}. No empirical range for
\nolinkurl{lambda} is asserted. Because the information tariff defines \nolinkurl{C=I(W;R)}, its 51,891
positive crossings are algebraically the same set as the 51,891
higher-information/higher-payoff module comparisons; they are not independent
evidence.

Equivalent symbolic quotients are canonicalized before distinct values are counted.
For the representation-information tariff the modal value is
\nolinkurl{1/(8-3*log2(3)) }\ensuremath{\approx}\nolinkurl{ 0.308155726702}, occurring 17,100 times.
The raw symbolic export contains 37 textual quotient forms; exact algebraic
normalization collapses these to 17 distinct numerical threshold values. Thus “17”
counts canonical values, whereas “37” counts pre-canonical symbolic spellings.

\FloatBarrier
\section*{S8. General boundary propositions}

\begin{enumerate}

\item \textbf{Refinement monotonicity.} If a deterministic representation \nolinkurl{r_f} strictly
   refines \nolinkurl{r_c} and both optimize over all deterministic latent-conditioned policies,
   then \nolinkurl{V*(r_f) >= V*(r_c)}. Proof: compose any policy on \nolinkurl{r_c} with the coarsening
   map from \nolinkurl{r_f}; the finer policy class contains an imitation of every coarser
   policy. For the broader comparison-of-experiments framework, see
   \href{https://doi.org/10.1214/aoms/1177729032}{Blackwell (1953)}, listed in the
   manuscript references. The argument here is the explicit deterministic
   policy-composition proof above, with no additional assumptions imported.

\item \textbf{Binary complement symmetry.} Suppose actions are binary, the decision-code
   library is closed under bitwise complement, both identity and bit-flip action codes
   are present, and payoff is symmetric zero-one correctness. Then each realized
   action map has at least two encodings. Paired fixed-background payoff contrasts
   have opposite signs while representation information is unchanged. The sign-pairing
   statement is not asserted outside this domain.

\item \textbf{Census non-invariance.} Let phenotype \nolinkurl{i} have encoding multiplicity \nolinkurl{m_i}. A
   census that includes all unordered encoded pairs contains
   \nolinkurl{sum_i choose(m_i,2)} within-phenotype pairs and
   \nolinkurl{sum_{i<j} m_i*m_j} between-phenotype pairs. Uniformly multiplying every \nolinkurl{m_i} by
   the same factor does \textbf{not} generally preserve counts or class shares because the
   two terms scale differently. If within-phenotype pairs are excluded and a class
   depends only on the phenotype pair, uniform duplication multiplies every class by
   the same squared factor and shares are invariant. Therefore every census must state
   both its pair universe and its counting measure.

\item \textbf{Information-tariff identity.} For \nolinkurl{C=I(W;R)}, a pair has a positive threshold
   exactly when its zero-cost payoff leader has higher retained world information.
   This follows directly from \nolinkurl{lambda*=delta_V/delta_C > 0}; it is an algebraic
   identity, not independent empirical support.

\end{enumerate}

These propositions are conditional on the policy and coding assumptions stated above.
The first, the explicitly conditional third, and the fourth have forms that apply
beyond the particular four-state counts. The second is exact only in the binary,
complement-closed, symmetric-payoff domain just stated. None is a biological theorem.

\FloatBarrier
\section*{S9. Tie and null structure}

All 1,267 module/environment optima are set-valued. Stable tasks have optimum-set
sizes 2, 4, or 8; two-task mixtures additionally have size 16; the all-task mixture
has sizes 2 or 4. No singleton optimum can exist because every realized action map
has a decision-complement/action-flip encoding partner. Neutral encodings are
retained in raw and aggregate artifacts; their existence is not interpreted as an
evolutionary result.

Likewise, the 792:792 strict representation-replacement count is forced by the
action-complement pairing. Each paired background preserves information order and
negates payoff difference. The existence of both strict directions is empirical to
the finite atlas; their exact equality is a model invariant.

Null and opposite-direction results are part of the main evidence: equal-information
comparisons, unequal-information payoff ties, strict higher-information/higher-payoff
comparisons, and tariff classes without positive crossings are not filtered.

\FloatBarrier
\section*{S10. Independent validation}

Fresh-context critical reviews accepted all three executions with zero findings.
Their reports record blind first-principles reconstruction, and repository checker
evidence is bound where it exists. The validated scope was:

\begin{itemize}

\item the complete sets of Stage 7.2A raw and aggregate results;

\item all 44,775 Stage 7.2B scientific rows;

\item all 916,265 Stage 7.2C scientific rows, symbolic threshold signs, tie sets, and
  aggregates.

\end{itemize}

The Stage 7.2B and 7.2C repository checker implementations and oracle evidence declare
no import of their production scientific cores. The Stage 7.2A review report records
a standalone blind reconstruction; its compact oracle evidence is separately bound.
Publication figures and robustness tables are generated from accepted saved
raw/aggregates by \nolinkurl{python -m fts_lab.publication.paper_001}.

\FloatBarrier
\section*{S11. Claim boundary}

Allowed:

\begin{quote}

Within the complete locked uniform four-state model, lower retained world
information can accompany higher expected task payoff, but the reverse relation
and ties also occur. Their zero-cost census changes with the comparison universe
and registered task mixture, while separately declared mechanism tariffs change
pairwise \nolinkurl{J_lambda} net-performance ordering.

\end{quote}

This is the content-addressed claim-ledger wording. In ordinary journal language,
“complete locked” means completely enumerated within the finite S2 domain, and
“registered task mixture” means one of the seven declared stationary objectives. It
does not imply public preregistration.

Not supported:

\begin{itemize}

\item \nolinkurl{I(W;R)} is truth simpliciter;

\item task payoff or \nolinkurl{J_lambda} is reproductive fitness;

\item pairwise order is invasion, fixation, extinction, or prevalence;

\item contextual locus contrasts are context-free causal allele effects;

\item project tariffs are empirical costs;

\item finite counts are probabilities over agents;

\item the model represents a named organism, human perception, or arbitrary ML systems;

\item the results prove FBT, ITP, consciousness, spacetime, ontology, or metaphysics.

\end{itemize}

\FloatBarrier
\section*{S12. AI-assisted workflow disclosure}

OpenAI Codex assisted with source organization, implementation, drafting, and
fresh-context review. The Human Principal Investigator approved the scientific
assumptions, byte-exact configurations, publication claim, and internal lifecycle.
Deterministic oracles and isolated verifier runs checked artifacts. The first author reviewed scientific decisions and interpretations and retains
responsibility for the manuscript. Internal AI-assisted checks are not external
human peer review. AI systems are not listed as authors.

\FloatBarrier
\section*{S13. Artifact map}

The accompanying \nolinkurl{PAPER-001-code-and-data.zip} is an article-scoped public package.
After extraction, its \nolinkurl{README.txt} and \nolinkurl{package_manifest.json} identify all files:

\begin{itemize}

\item current editable manuscripts and supplement: \nolinkurl{publications/paper-001/};

\item formal definitions and exact fixtures: \nolinkurl{specs/integrated/};

\item unchanged exact configurations: \nolinkurl{experiments/configs/};

\item unchanged raw and derived results: \nolinkurl{results/raw/} and \nolinkurl{results/derived/};

\item original computation manifests: \nolinkurl{experiments/manifests/};

\item figures: \nolinkurl{publications/paper-001/figures/} and \nolinkurl{figures_ru/};

\item figure and supplementary-table inputs: \nolinkurl{publications/paper-001/source_tables/};

\item figure checksums and input bindings: \nolinkurl{publications/paper-001/figure_manifest.json};

\item claim-to-evidence trace: \nolinkurl{publications/paper-001/claim_evidence_ledger.csv};

\item scientific core, independent checkers, and saved-data figure builder: \nolinkurl{src/fts_lab/};

\item portable verification and reproduction commands: \nolinkurl{README.txt}, \nolinkurl{verify_package.py};

\item pinned verification dependencies: \nolinkurl{requirements-verification.txt};

\item original project dependency files: \nolinkurl{environment/} (historical provenance).

\end{itemize}

Machine-readable table filenames retain their original identifiers. Current
supplementary table numbers map to them as follows:

\begin{itemize}

\item Table S3: \nolinkurl{table_s5_information_decomposition_witness.csv};

\item Table S8: \nolinkurl{table_s6_denominator_map.csv};

\item Table S11: \nolinkurl{table_s1_representation_once_census.csv};

\item Table S12: \nolinkurl{table_s2_nested_refinement_control.csv};

\item Table S14: \nolinkurl{table_s3_one_coordinate_diagnostics.csv};

\item Table S15: \nolinkurl{table_s7_worked_order_examples.csv};

\item Table S17: \nolinkurl{table_s4_threshold_distributions.csv}.

\end{itemize}

Historical review documents named by original manifests or the claim ledger are
not distributed here. They record internal AI-assisted checking, not external
human peer review. The public verification commands do not require those files.
Original computational provenance, including generic workspace paths and Git
identifiers, is retained to preserve raw-file identity. The original non-executable
configuration flags remain unchanged; see README.txt for the public verification
workflow. Primary-source papers and private correspondence are not included.

\end{document}